\documentclass{article}
\usepackage{arxiv}

\usepackage{graphicx,amsthm} 
\usepackage{subfigure}
\usepackage{wrapfig}
\usepackage{sidecap}
\usepackage{multicol}

\usepackage{algorithmic}
\usepackage[ruled,vlined,noend,linesnumbered]{algorithm2e}
\usepackage{bm}
\usepackage{amsmath}
\usepackage{amssymb}
\usepackage{amsfonts}
\usepackage{latexsym}
\usepackage{multirow,tikz,amsmath,comment}
\usepackage{enumitem}
\usepackage[utf8]{inputenc} 
\usepackage[T1]{fontenc}    
\usepackage{hyperref}       
\usepackage{url}            
\usepackage{booktabs}       
\usepackage{amsfonts}       
\usepackage{nicefrac}       
\usepackage{microtype}      
\usepackage{sidecap}
\usepackage{natbib}
\usepackage{mathtools}
\usepackage{txfonts}

\usepackage{multicol}
\usepackage{threeparttable}

\usepackage{thmtools,thm-restate}
\newtheorem{theorem}{Theorem}

\declaretheorem[sibling=theorem]{lemma}
\newtheorem{definition}{Definition}

\newtheorem{condition}{Condition}
\newtheorem{example}{Example}

\newcommand{\lacl}{\color{brown}}

\newcommand{\setL}{\mathbf{L}}
\newcommand{\setX}{\mathbf{X}}
\newcommand{\setV}{\mathbf{V}}
\newcommand{\setA}{\mathbf{A}}
\newcommand{\setB}{\mathbf{B}}
\newcommand{\setC}{\mathbf{C}}

\newcommand{\setP}{\mathbf{P}}
\newcommand{\setS}{\mathbf{S}}
\newcommand{\graph}{\mathcal{G}}
\newcommand{\graphp}{\mathcal{G'}} 
\newcommand{\latents}{\setL_{\graph}}

\newcommand{\measures}{\setX_{\graph}}
\newcommand{\measuresp}{\setX_{\graphp}}
\newcommand{\variables}{\setV_{\graph}}

\newcommand{\grandparentsp}{Gp_{\graphp}}
\newcommand{\parents}{Pa_{\graph}}
\newcommand{\parentsp}{Pa_{\graphp}} 

\newcommand{\childrenp}{Ch_{\graphp}}

\newcommand{\siblingsp}{Sib_{\graphp}}
\newcommand{\purechildren}{PCh_{\graph}}
\newcommand{\purechildrenp}{PCh_{\graphp}}
\newcommand{\puredescendant}{PDe_{\graph}}

\newcommand{\measured}{\mathcal{M}_{\graph}}
\newcommand{\measuredp}{\mathcal{M}_{\graphp}}

\SetNlSty{}{}{}
\let\oldnl\nl
\newcommand\nonl{%
  \renewcommand{\nl}{\let\nl\oldnl}}
  
\SetCommentSty{mycommfont}

\title{Latent Hierarchical Causal Structure Discovery with Rank Constraints}

\author{ \hspace{-3.5mm}
\textbf{Biwei Huang} $^{*1}$ \textbf{Charles Low} \thanks{These authors contributed equally to this work. Accepted at 36th Conference on Neural Information Processing Systems (NeurIPS 2022).} $^{1}$, \textbf{Feng Xie}$^{3}$, \textbf{Clark Glymour}$^{1}$, \textbf{Kun Zhang}$^{1,2}$\\ 
$^1$ Carnegie Mellon University\\
$^2$ Mohamed bin Zayed University of Artificial Intelligence\\
$^3$ Beijing Technology and Business University, China\\
\texttt{\{bwei.huang, charleslow88, xiefeng009\}@gmail.com,}\\
\texttt{cg09@andrew.cmu.edu, kunz1@cmu.edu}
}

\begin{document}

\maketitle
\vspace{-2mm}
\begin{abstract}
 Most causal discovery procedures assume that there are no latent confounders in the system, which is often violated in real-world problems. In this paper, we consider a challenging scenario for causal structure identification, where some variables are latent and they form a hierarchical graph structure to generate the measured variables; the children of latent variables may still be latent and only leaf nodes are measured, and moreover, there can be multiple paths between every pair of variables (i.e., it is beyond tree structure). We propose an estimation procedure that can efficiently locate latent variables, determine their cardinalities, and identify the latent hierarchical structure, by leveraging rank deficiency constraints over the measured variables. We show that the proposed algorithm can find the correct Markov equivalence class of the whole graph asymptotically under proper restrictions on the graph structure.
\end{abstract} 

\section{Introduction}

In many cases, the common assumption in causal discovery algorithms--no latent confounders--may not hold. For example, in complex systems, it is usually hard to enumerate and measure all task-related variables, so there may exist latent variables that influence multiple measured variables, the ignorance of which may introduce spurious correlations among measured variables. Much effort has been made to handle the confounding problem in causal structure learning. One research line considers the causal structure over measured variables, including FCI and its variants \citep{spirtes2000causation,Pearl:2000:CMR:331969,colombo2012learning,Akbari2021}, matrix decomposition-based approaches~\citep{RankSparsity_11, RankSparsity_12,frot2019robust}, and over-complete ICA-based ones~\citep{hoyer2008estimation,salehkaleybar2020learning}. 

Another line focuses on identifying the causal structure among latent variables, including Tetrad condition-based approaches~\citep{Silva-linearlvModel, Kummerfeld2016}, high-order moments-based ones~\citep{shimizu2009estimation,cai2019triad,xie2020generalized,adams2021identification,chen2022identification}, matrix decomposition-based approach \citet{anandkumar2013learning}, copula model-based approach~\citep{cui2018learning}, mixture oracles-based approach~\citep{kivva2021learning}, and multiple domains-based approach~\citep{zeng2021causal}. Moreover, regarding the scenario of latent hierarchical structures, previous work along this line assumes a tree structure~\citep{Pearl1988,zhang2004hierarchical,choi2011latenttree,drton2017marginal}, where there is one and only one undirected path between every pair of variables. This assumption is rather restrictive and the structure in real-world problems could be more complex--beyond a tree.

\begin{figure}[htp!]
\centering
    \subfigure[]{
	\begin{tikzpicture}[scale=.7, line width=0.4pt, inner sep=0.1mm, shorten >=.1pt, shorten <=.1pt]
		\draw (0, 2.5) node(L1)  {{\footnotesize\lacl\,{$L_1$}\,}};
		\draw (-1.5, 1) node(L2) {{\footnotesize\lacl\,{$L_2$}\,}};
		\draw (-0.5, 1) node(L3)  {{\footnotesize\lacl\,$L_3$\,}};
		\draw (1.8, 1) node(L4)  {{\footnotesize\lacl\,$L_4$\,}};
		\draw (2.8, 1) node(L5)  {{\footnotesize\lacl\,$L_5$\,}};
		
		\draw (-3, -0.5) node(L6)  {{\footnotesize\lacl\,$L_6$\,}};
		\draw (-2, -0.5) node(L7)  {{\footnotesize\lacl\,$L_7$\,}};
		\draw (-1.2, -0.5) node(L8)  {{\footnotesize\lacl\,$L_8$\,}};
		\draw (0, -0.5) node(L9)  {{\footnotesize\lacl\,$L_9$\,}};
		\draw (0.8, -0.5) node(L10)  {{\footnotesize\lacl\,$L_{10}$\,}};
		
		\draw (1.75, -0.5) node(X12)  {{\footnotesize\,$X_{12}$\,}};
		\draw (2.5, -0.5) node(X13)  {{\footnotesize\,$X_{13}$\,}};
		\draw (3.25, -0.5) node(X14)  {{\footnotesize\,$X_{14}$\,}};
		\draw (4.05, -0.5) node(X15)  {{\footnotesize\,$X_{15}$\,}};
		\draw (4.8, -0.5) node(X16)  {{\footnotesize\,$X_{16}$\,}};
		
		\draw (-4, -2) node(X1)  {{\footnotesize\,$X_1$\,}};
		\draw (-3.5, -2) node(X2)  {{\footnotesize\,$X_2$\,}};
		\draw (-3, -2) node(X3)  {{\footnotesize\,$X_3$\,}};
		
		\draw (-2.3, -2) node(X4)  {{\footnotesize\,$X_4$\,}};
		\draw (-1.7, -2) node(X5)  {{\footnotesize\,$X_5$\,}};
		\draw (-1.1, -2) node(X6)  {{\footnotesize\,$X_6$\,}};
		\draw (-0.5, -2) node(X7)  {{\footnotesize\,$X_7$\,}};
		
		\draw (0.15, -2) node(X8)  {{\footnotesize\,$X_8$\,}};
		\draw (0.8, -2) node(X9)  {{\footnotesize\,$X_9$\,}};
		\draw (1.45, -2) node(X10)  {{\footnotesize\,$X_{10}$\,}};
		\draw (2.2, -2) node(X11)  {{\footnotesize\,$X_{11}$\,}};
		
	   \draw[-latex] (L1) -- (L2);
	   \draw[-latex] (L1) -- (L3);
	   \draw[-latex] (L1) -- (L4);
	   \draw[-latex] (L1) -- (L5);
	   \draw[-latex] (L2) -- (L6);
	   \draw[-latex] (L2) -- (L7);
	   \draw[-latex] (L2) -- (L8);
	   \draw[-latex] (L2) -- (L9);
	   \draw[-latex] (L2) -- (L10);
	   \draw[-latex] (L3) -- (L7);
	   \draw[-latex] (L3) -- (L8);
	   \draw[-latex] (L3) -- (L9);
	   \draw[-latex] (L4) -- (X12);
	   \draw[-latex] (L4) -- (X13);
	   \draw[-latex] (L4) -- (X14);
	   \draw[-latex] (L4) -- (X15);
	   \draw[-latex] (L4) -- (X16);
	   \draw[-latex] (L5) -- (X12);
	   \draw[-latex] (L5) -- (X13);
	   \draw[-latex] (L5) -- (X14);
	   \draw[-latex] (L5) -- (X15);
	   \draw[-latex] (L5) -- (X16);
	   
	   \draw[-latex] (L6) -- (X1);
	   \draw[-latex] (L6) -- (X2);
	   \draw[-latex] (L6) -- (X3);
	   \draw[-latex] (L7) -- (X4);
	   \draw[-latex] (L7) -- (X5);
	   \draw[-latex] (L7) -- (X6);
	   \draw[-latex] (L7) -- (X7);
	   \draw[-latex] (L8) -- (X6);
	   \draw[-latex] (L8) -- (X7);
	   \draw[-latex] (L9) -- (X8);
	   \draw[-latex] (L9) -- (X9);
	   \draw[-latex] (L9) -- (X10);
	   \draw[-latex] (L9) -- (X11);
	   \draw[-latex] (L10) -- (X8);
	   \draw[-latex] (L10) -- (X9);
	   \draw[-latex] (L10) -- (X10);
	   \draw[-latex] (L10) -- (X11);
	   \label{fig:hierarchical-a}
	\end{tikzpicture}
    }
    \subfigure[]{
    \begin{tikzpicture}[scale=.72, line width=0.4pt, inner sep=0.2mm, shorten >=.1pt, shorten <=.1pt]
    \draw (0, 0) node(X6)  {{\footnotesize\,$X_6$\,}};
	\draw (0, -0.7) node(X7) {{\footnotesize\,$X_7$\,}};
	\draw (0, -1.4) node(X8)  {{\footnotesize\,$X_8$\,}};
	\draw (0, -2.1) node(X9)  {{\footnotesize\,$X_9$\,}};
	
	\draw (1.2, -0.5) node(L4)  {{\footnotesize\lacl\,$L_4$\,}};
	\draw (1.2, -1.5) node(L5)  {{\footnotesize\lacl\,$L_5$\,}};
	
	\draw (2.4, 0) node(L2)  {{\footnotesize\lacl\,$L_2$\,}};
	\draw (2.4, -1) node(L1) {{\footnotesize\lacl\,$L_1$\,}};
	\draw (2.4, -2) node(L3)  {{\footnotesize\lacl\,$L_3$\,}};
	\draw (3.6, 0.6) node(X4)  {{\footnotesize\,$X_4$\,}};
	\draw (3.6, -2.6) node(X5) {{\footnotesize\,$X_5$\,}};
	
	\draw (3.6, 0) node(L6)  {{\footnotesize\lacl\,$L_6$\,}};
	\draw (3.6, -0.8) node(X0) {{\footnotesize\,$X_0$\,}};
	\draw (3.6, -1.2) node(X1) {{\footnotesize\,$X_1$\,}};
	\draw (3.6, -2) node(L7)  {{\footnotesize\lacl\,$L_7$\,}};
	
	\draw (4.8, 0.3) node(X10)  {{\footnotesize\,$X_{10}$\,}};
	\draw (4.8, -0.3) node(X11)  {{\footnotesize\,$X_{11}$\,}};
	\draw (4.8, -1) node(X12)  {{\footnotesize\,$X_{12}$\,}};
	\draw (4.8, -1.7) node(X3)  {{\footnotesize\,$X_{3}$\,}};
	\draw (4.8, -2.3) node(X2)  {{\footnotesize\,$X_{2}$\,}};
	
	\draw[-latex] (L4) -- (X6);
	\draw[-latex] (L4) -- (X7);
	\draw[-latex] (L4) -- (X8);
	\draw[-latex] (L4) -- (X9);
	\draw[-latex] (L5) -- (X6);
	\draw[-latex] (L5) -- (X7);
	\draw[-latex] (L5) -- (X8);
	\draw[-latex] (L5) -- (X9);
	
	\draw[-latex] (L2) -- (L4);
	\draw[-latex] (L2) -- (L5);
	\draw[-latex] (L3) -- (L4);
	\draw[-latex] (L3) -- (L5);
	
	\draw[-latex] (L2) -- (X4);
	\draw[-latex] (L1) -- (L2);
	\draw[-latex] (L1) -- (L3);
	\draw[-latex] (L3) -- (X5);
	
	\draw[-latex] (L2) -- (L6);
	\draw[-latex] (L1) -- (X0);
	\draw[-latex] (L1) -- (X1);
	\draw[-latex] (L3) -- (L7);
	
	\draw[-latex] (L6) -- (X10);
	\draw[-latex] (L6) -- (X11);
	\draw[-latex] (L6) -- (X12);
	\draw[-latex] (L7) -- (X12);
	\draw[-latex] (L7) -- (X3);
	\draw[-latex] (L7) -- (X2);
	\label{fig:hierarchical-b}
	\end{tikzpicture}
    }
\caption{Example hierarchical graphs that our method can handle,where $X_i$ are measured variables and $L_i$ are latent variables.}
\label{fig:hierarchical}
\end{figure}
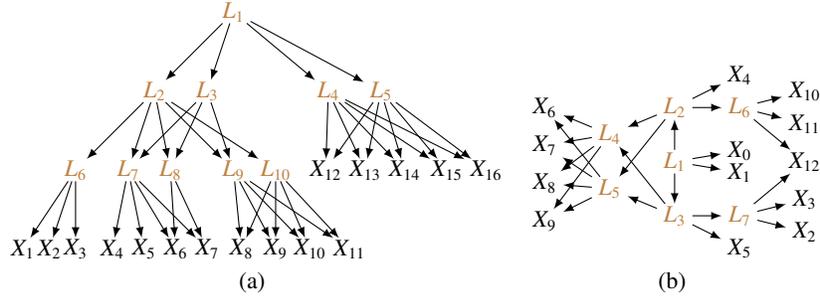

In this paper, we consider a more challenging scenario where latent causal variables form a hierarchical graph structure to generate measured variables---the children of latent variables may still be latent and only the leaf nodes are measured, and moreover, there can be multiple paths between every pair of variables (see the example hierarchical graphs in Figure \ref{fig:hierarchical}). 
We aim to find out identifiability conditions of the hierarchical structure that are as mild as possible, and meanwhile, develop an efficient algorithm with theoretical guarantees to answer the following questions. (1) How can we locate latent parents for both measured and latent variables, as well as determining the cardinality of the latent parents, by only providing the leaf nodes? (2) How can we identify the causal relationships among latent variables and those from latent variables to measured variables?

Interestingly, we can answer these questions by properly making use of rank deficiency constraints; finding and leveraging rank properties in specific ways enable us to identify the Markov equivalence class of the whole graph, under appropriate conditions. Our contribution is mainly two-fold:
\begin{itemize}[leftmargin=15pt,itemsep=0pt,topsep=0pt]
    \item We propose a structure identification algorithm that can efficiently locate  latent variables (including their cardinalities) and identify the latent hierarchical structure, by leveraging the rank deficiency. 
    \item We show that the proposed algorithm can find the correct graph asymptotically under mild restrictions of the graph structure. Roughly speaking, we show that it is sufficient to have $k+1$ pure children (which can be latent), as well as another $k+1$ neighbors, to identify the latent variable set with size $k$ (see the detailed conditions in Definition \ref{Def:latent atomic cover} and Condition \ref{Condition:R2H}).
\end{itemize}

It is worth mentioning that rank constraints have been used in previous methods \citep{Silva-linearlvModel, Kummerfeld2016}, but they assume that each latent variable has three measured ones as children and each measured variable has only one latent parent. There are also other methods for latent structure learning; for instance, \citet{anandkumar2013learning}, which uses matrix decomposition, needs $3k$ measured children, and the GIN-based method \citep{xie2020generalized}, which makes use of high-order statistics, needs $2k$ measured children. However, all those developments require that every latent variable should have measured variables as children. Very recently, \citet{Hierarchical_Xie22} proposes an approach for latent hierarchical structure by leveraging the GIN condition under linear non-Gaussian models, but it assumes that each variable has only one parent, where both figures in Figure \ref{fig:hierarchical} do not satisfy.

This paper is organized as follows. In Section \ref{Sec:def}, we give formal definitions of the latent hierarchical causal model under investigation and give conditions that are essential to the identifiability of the graph structure. In Section \ref{Sec:Alg}, we propose an efficient algorithm that makes use of rank-deficiency constraints to identify the latent hierarchical structure. Moreover, we show theoretically in Section \ref{Sec:theory} that the proposed algorithm outputs the correct Markov equivalence class of the whole graph asymptotically. 
In Section \ref{Sec:exp}, we empirically validate the proposed approach on synthetic data.  Notations for graphical representations that are used in the paper are provided in Table \ref{Table:notation}.

\begin{table}[hpt!]
\resizebox{\textwidth}{!}{%
\begin{tabular}{|l|l|l|}
\hline
Pa: ~~~Parents           & Sib: Siblings  & $\mathbf{V}_{\graph}$: All variables in graph $\graph$ \\ \hline
PCh: Pure children & $\mathcal{M}$: ~Measured pure descendants & $\mathbf{X}$: ~A set of measured variables \\ \hline
PDe: Pure descendants & $\mathbf{X}_{\graph}$: All measured variables in graph $\graph$   & $\mathbf{L}$: ~A set of latent variables         \\ \hline
Gp: ~~Grandparents      &    $\mathbf{L}_{\graph}$: ~All latent variables in graph $\graph$         &     $\mathbf{V}$: ~A set of variables                                      \\ \hline
\end{tabular}%
}
\caption{Graphical notations used in the paper.}
\label{Table:notation}
\end{table}

\section{Latent Hierarchical Causal Model}\label{Sec:def}

In this paper, we focus on latent hierarchical causal model with graph structure $\graph$, where both measured variables $\mathbf{X}_{\graph} = \{X_1, \cdots, X_m\}$ and latent variables $\mathbf{L}_{\graph} = \{L_1, \cdots, L_n\}$ are generated by their latent parents in a directed acyclic graph (DAG) with linear relationships:
\begin{equation}
    \begin{array}{l}
          X_i = \sum \nolimits_{L_j \in Pa(X_i)} b_{ij}L_j + \varepsilon_{X_i},\\
          L_j = \sum \nolimits_{L_k \in Pa(L_j)} c_{jk}L_k + \varepsilon_{L_j},
    \end{array}
\label{Eq.model}
\end{equation}
where $b_{ij}$ and $c_{jk}$ are the causal strength from $L_j$ to $X_i$ and from $L_k$ to $L_j$, respectively, and $\varepsilon_{X_i}$ and $\varepsilon_{L_j}$ are noise terms that are independent of each other.  Without loss of generality, we assume that all variables in $\mathbf{X}_{\graph}$ and $\mathbf{L}_{\graph}$ have zero mean.

Below, we first give the general definition of a linear latent hierarchical graphical model in Definition \ref{Def:L2H}. Then we give more detailed conditions on the graph structure (Definitions \ref{Def:pure}-\ref{Def:latent atomic cover}) that are essential to formalize the identifiability condition of the latent hierarchical structure.

\begin{definition}[Linear Latent Hierarchical (L$^2$H)  Model]
A graphical model, with its graph $\graph = (\mathbf{V}_{\graph}, \mathbf{E}_{\graph})$, is a linear latent hierarchical model if:
\begin{enumerate}[leftmargin=15pt,align=left,itemsep=0pt,topsep=0pt]
    \item $\mathbf{V}_{\graph} = \mathbf{X}_{\graph} \bigcup \mathbf{L}_{\graph}$, where $\mathbf{X}_{\graph}$ is the set of measured variables and $\mathbf{L}_{\graph}$ is the set of latent variables,
    \item there is at least one undirected path between every pair of variables, and
    \item each variable in $\mathbf{X}_{\graph}$ and $\mathbf{L}_{\graph}$ are generated by the structural equation models in Eq. \ref{Eq.model}.
\end{enumerate}
\label{Def:L2H}
\end{definition}

Generally, without further constraints, the causal structure of the L$^2$H model is hard to be identified. It has been shown that if the underlying graph structure satisfies a tree \citep{Pearl1988}, then the structure is identifiable. However, this structural constraint may be too strong to hold in many real-world problems. In this paper, we give sufficient conditions that are much milder than previous ones, as well as an efficient search algorithm, for the identifiability of the causal structure. 

We now give the corresponding definitions, including \textit{pure children}, \textit{pure descendants}, and \textit{effective cardinality}, that will be used in the identifiability condition, together with illustrative examples. 

\begin{definition}[Pure Children]
Variables $\setV$ are pure children of a set of latent variables $\setL$ in a graph $\graph$, if $\parents(\setV) = \setL$ and $\setV \cap \setL=\emptyset$. That is, $\setV$ have no other parents than $\setL$. We denote the pure children of $\setL$ by $\purechildren(\setL)$. 
\label{Def:pure}
\end{definition}

Accordingly, \textit{Pure Descendants} of a set of latent variables $\setL$ are defined as all recursive pure children of $\setL$ (including $\purechildren(\setL), \purechildren(\purechildren(\setL))$, etc.), denoted by $\puredescendant(\setL)$. Furthermore, measured variables that are pure descendent of $\setL$ are called \textit{Measured Pure Descendants}, denoted by $\measured(\setL)$.

\begin{example}
In Figure \ref{fig:hierarchical-a}, the pure children of $\{L_2,L_3\}$ are $\{L_6,\cdots,L_{10}\}$, its pure descendants  are $\{L_6,\cdots,L_{10},X_1,\cdots,X_{11}\}$, and its measured pure descendants are $\{X_1,\cdots,X_{11}\}$.
\end{example}

\begin{definition}[Effective Cardinality]
For a set of latent variables $\setL$, denote by $\setC$ the largest subset of $\purechildren(\setL)$ such that there is no subset $\setC' \subseteq \ \setC$ satisfying $|\setC'| > |\parents(\setC')|$ and $|\parents(\setC')|<|\setL|$.
Then, the effective cardinality of $\setL$'s pure children is $|\setC|$.
\end{definition}

In the case when $\setL$ and its pure children are fully connected, the effective cardinality is just the cardinality of the pure children of $\setL$. However, for Figure \ref{fig:hierarchical-a}, the effective cardinality of the pure children of $\{L_7,L_{8}\}$ is 3, because here the largest subset that satisfies the condition is $\{ X_5, X_6, X_7\}$.

We further define \textit{latent atomic cover} that constrains the number of pure children and neighbours for latent variables, which are essential for structural identifiability.	
\begin{definition}
[Latent Atomic Cover] Let $\setL = \{ L_1, ..., L_k\}$ be a set of latent variables in graph $\graph$, with $|\setL| = k$. We say that $\setL$ is a latent atomic cover if the following conditions are met:
\begin{enumerate}[leftmargin=7pt,align=left,itemsep=0pt,topsep=0pt]
    \item there exists a subset of pure children $\setC' \subseteq PCh_{\graph}(\setL)$ with effective cardinality $\geq k+1$;
    \item there exists a neighbour set $\setB$ to $\setL$ s.t. $\setB \bigcap \setC' = \emptyset$ and $|\setB| = k+1$;  
    \item there does not exist a partition of $\setL = \setL_1 \cup \setL_2$, so that both $\setL_1,\setL_2$ satisfy conditions 1 and 2 and $\{PCh_{\graph}(\setL_1) \cup PCh_{\graph}(\setL_2)\} \backslash \setL = PCh_{\graph}(\setL)$.
\end{enumerate}
\label{Def:latent atomic cover}
\end{definition}

\begin{example}
  In Figure \ref{fig:hierarchical-a}, $\mathbf{L} =\{L_2,L_3\}$ is a latent atomic cover with $k=2$, because (1) there exists a subset of pure children $\setC' = \{L_6,L_7,L_8\}$ with effective cardinality $3= k+1$, (2) there exists a neighbor set $\setB = \{L_1,L_9,L_{10}\}$, s.t. $\setB \cap \setC' = \emptyset$ and $|\setB| = 3 = k+1$, and (3) neither $\{L_2\}$ or $\{L_3\}$ satisfies the above two conditions.
\end{example}

We now give the conditions for structural identifiability from measured variables $\setX_{\graph}$ alone, including those on the structural constraints (Condition \ref{def-rllhm}) and rank faithfulness assumption (Condition \ref{Con:rankfaithful}).

\begin{condition}[Irreducible Linear Latent Hierarchical (IL$^2$H) Graph]
\label{def-rllhm}
An L$^2$H graph $\graph$ is an IL$^2$H graph if 
\begin{enumerate}[leftmargin=*,align=left,itemsep=0pt,topsep=0pt]
    \item every latent variable $L \in \mathbf{L}$ in $\graph$ belongs to at least one latent atomic cover, 
    \item for any pair of latent atomic covers $(\setL_A, \setL_B)$, if $\puredescendant(\setL_A) \bigcap \puredescendant(\setL_B) \neq \emptyset$, then either (a) $\setL_A \subset \setL_B$ or (b) $\setL_A \subset \puredescendant(\setL_B)$ or (c) $\setL_B \subset \setL_A$ or (d) $\setL_B \subset \puredescendant(\setL_A)$, and
    \item for any three latent atomic covers $\setL_A, \setL_B, \setL_C$, if the causal structure satisfies $\setL_A \rightarrow \setL_B \rightarrow \setL_C$ or $\setL_A \leftarrow \setL_B \rightarrow \setL_C$, and $|\setL_B|=k$, then $\setL_B$ has $2k$ neighbors, except for $\setL_A, \setL_C$ and the parents in the v structure where $\setL_B$ is a collider.
\end{enumerate}
\label{Condition:R2H}
\end{condition}

Next, we give the faithfulness assumption, which holds for generic covariance matrices consistent with $\graph$ ~\citep{spirtes2013calculation-t-separation}.
\begin{condition}[Rank Faithfulness]
A probability distribution $P$ is rank faithful to a DAG $\mathcal{G}$ if every rank constraint on a sub-covariance matrix that holds in $P$ is entailed by every linear structural model with respect to $\mathcal{G}$.
\label{Con:rankfaithful}
\end{condition}

We will show in Section \ref{Sec:theory} that if the underlying latent hierarchical graph satisfies an IL$^2$H graph and the rank faithfulness holds, then the location and cardinality of latent variables, and the causal structure among latent atomic covers and that from latent atomic covers to measured variables, are identifiable with appropriate search procedures.  Below, let us first present the identification procedure.

\section{Structure Identification with Rank-Deficiency Constraints}\label{Sec:Alg}

We propose an identification algorithm (Algorithm \ref{overallAlg}) to identify the structure of IL$^2$H graphs, by leveraging rank-deficiency constraints of measured variables. In particular, the algorithm includes three phases: finding causal clusters and assigning latent atomic covers in a greedy manner (``\textit{findCausalClusters}"), refining incorrect clusters and covers due to the greedy search (``\textit{refineClusters}"), and refining edges and finding v structures (``\textit{refineEdges}").
\begin{algorithm}[!htp]
\SetKwInOut{Input}{Input}
\SetKwInOut{Output}{Output}
\SetKwFunction{overallAlg}{\texttt{overallAlg}}
\Input{Date from a set of measured variables $\measures$}
\Output{Markov equivalence class $\graph'$}%

$\graph'$ = findCausalClusters ($\measures$) \tcp*{\footnotesize find clusters and assign latent covers greedily}

$\graph'$ = refineClusters ($\graph'$) \tcp*{\footnotesize refine incorrect clusters and covers from greedy search}

$\graph'$ = refineEdges ($\graph'$) \tcp*{\footnotesize refine some edges and find v structures}
\caption{Latent Hierarchical Causal Structure Discovery}
\label{overallAlg}
\end{algorithm}

Before describing the identification algorithm, we first give a theorem that relates the graphical structure of an IL$^2$H graph to the rank constraints over the covariance matrix of measured variables.

\begin{theorem}[Graphical Implication of Rank Constraints in IL$^2$H Graphs]
\label{dsep-theorem}
Suppose $\mathcal{G}$ satisfies an IL$^2$H graph. Under the rank faithfulness assumption, the cross-covariance matrix $\Sigma_{\setX_A,\setX_B}$ over measured variables $\setX_A$ and $\setX_B$ in $\mathcal{G}$ (with $|\setX_A|, |\setX_B|>r$) has rank $r$, if and only if there exists a subset of latent variables $\setL$ with $|\setL| = r$ such that $\setL$ d-separates $\setX_A$ from $\setX_B$, and there is no $\mathbf{L}'$ with $|\mathbf{L}'| < |\setL|$ that d-separates $\setX_A$ from $\setX_B$. That is,
$$rank(\Sigma_{\setX_A,\setX_B}) = min \{ |\setL|: \setL \text{ d-separates } \setX_A \text{ from } \setX_B \}.$$
\end{theorem}
For instance, in Figure \ref{fig:hierarchical-a}, $rank(\Sigma_{\{X_1,X_2\}, \{X_3,X_4\}})=1$, because $L_6$ d-separates $\{X_1,X_2\}$ from $\{X_3,X_4\}$ with $|L_6|=1$. 

\paragraph{Rank Test.} We test rank deficiency by leveraging canonical correlations \citep{ranktest}. Specifically, the number of non-zero canonical correlations between two random vectors is equal to the rank. Denote by $\alpha_i$ the $i$-th canonical correlation coefficient between $\setX_A$ and $\setX_B$. Then under the null hypothesis that $\texttt{rank}(\Sigma_{\setX_A,\setX_B}) \leq r$, the statistics $$-(N\!-\!(p+q+3)/2) \sum_{i=r+1}^{\max(p,q)} \log(1-\alpha_i^2)$$ is approximately $\chi^2$ distributed with $(p-r)(q-r)$ degrees of freedom, where $p=|\setX_A|$, $q=|\setX_B|$, and $N$ is the sample size. 

We further show that for any subset of latent variables in an IL$^2$H graph, we can use the measured variables as surrogates to estimate the rank, as indicated in the following theorem. 
\begin{theorem}[Measurement as a surrogate]
 Suppose $\graph$ is an IL$^2$H graph. Denote by $\setA, \setB \subseteq \setV_{\graph}$ two subsets of variables in $\graph$, with $\setA \cap \setB = \emptyset$. Furthermore, denote by $\setX_A$ the set of measured variables that are d-separated by $\setA$ from all other measures, and by $\setX_B$ the set of measured variables that are d-separated by $\setB$ from all other measures. Then $\texttt{rank}(\Sigma_{\setA,\setB}) = \texttt{rank}(\Sigma_{\setX_A,\setX_B}).$
 \label{The:surrogate}
\end{theorem}
A special case when Theorem 2 holds is that $\setX_A$ and $\setX_B$ are the measured pure descendants of $\setA$ and $\setB$, respectively. Note that Theorem \ref{dsep-theorem} is a special case of Theorem 2.8 in \citet{tseparation} when applied to IL$^2$H graphs. Different from the setting in \citet{tseparation} where access to the full covariance matrix $\Sigma_{\variables,\variables}$ is assumed, we only have access to the covariance matrix $\Sigma_{\measures,\measures}$ over the measured variables $\measures$, which we will use to infer the causal structure over the entire graph $\graph$. For this reason, although we can infer the number of latent variables that d-separate any two sets of measured variables $\setX_A, \setX_B$, we cannot directly know the exact location of these variables in the graph. Fortunately, the structure constraints of the IL$^2$H graph and Theorem \ref{The:surrogate} will allow us to reconstruct the graph with certain search procedures, as shown below.

\subsection{Phase I: Finding Causal Clusters}
We start to discover clusters in a recursive and greedy manner, by performing rank-deficiency tests over measured variables. We denote by $\mathcal{S}$ a set of active variables that is under investigation; $\mathcal{S}$ is set to $\measures$ initially and will be updated to include latent atomic covers when rank deficiency is discovered. Below, we first give the definition about set size and the definition of \textit{atomic rank-deficiency set}, which will be used in Rule 1 and Algorithm \ref{findCausalClusters}.
\begin{definition}[Set Size]
 Suppose $\mathcal{S}$ is a set. We denote by $|\mathcal{S}|$ the cardinality of $\mathcal{S}$; that is, $|\mathcal{S}|$ is the number of elements $S_i$ in $\mathcal{S}$, for $S_i \in \mathcal{S}$. We denote by $\left\Vert \mathcal{S}\right\Vert$ the number of variables in $\mathcal{S}$, where $\left\Vert \mathcal{S}\right\Vert = |\bigcup_i S_i|$, for $S_i \in \mathcal{S}$.
\end{definition}
For example, suppose $\mathcal{S} = \big\{\{L_1\},\{L_6\},\{L_7,L_8\},\{L_9,L_{10}\}\big\}$. Then $|\mathcal{S}| = 4$ and $\left\Vert \mathcal{S}\right\Vert = 6$.

\begin{definition}[Atomic Rank-Deficiency Set]
 Given a graph $\graphp$. Denote by $\mathcal{S} \subseteq V_{\graphp}$ an active set of variables that is under investigation. Let $\setA \subset \mathcal{S}$ and $\setB = \mathcal{S} \backslash \setA$. Denote by $\setX_A = \measuredp(\setA)$ and $\setX_B = \measuredp(\setB)$ the measured pure descendants of $\setA$ and $\setB$, respectively. If (1) $\Sigma_{\setA, \setB}$ is rank deficient, i.e., $\texttt{rank}{(\Sigma_{\setA, \setB})} < min\{\left\Vert \setA \right\Vert, \left\Vert \setB \right\Vert\}$), and (2) no proper subset of $\setA$ is rank deficient, then $\setA$ is called an atomic rank-deficient set, and $\Sigma_{\setA, \setB}$ can be estimated by $\Sigma_{\setX_A, \setX_B}$. 
\end{definition}

For example, in Figure \ref{fig:hierarchical-a}, suppose now we have the measured variables $\setX_{\graph}$ under investigation, so the active variable set $\mathcal{S} = \setX_{\graph}$. $\setA = \{X_1, X_2\} \subset \mathcal{S}$ forms an atomic rank-deficiency set because the cross-covariance matrix of $\setX_A$ against all other measures has rank $1$ (i.e., rank deficient). According to Theorem \ref{dsep-theorem}, the rank deficiency occurs because the latent parent $L_6$ of $\{X_1, X_2\}$ d-separates them from all other measures. This naturally leads to the following rule that assigns a latent atomic cover over the rank-deficiency set:

\begin{itemize}[leftmargin=30pt,align=left,itemsep=0pt,topsep=0pt]
    \item[\textbf{Rule 1:}] If $\setA$ is a rank-deficiency set with $\texttt{rank}(\Sigma_{\setX_A, \setX_B}) = k$, then assign a latent atomic cover $\setL$ of size $k$ as the parent of every variable $A_i \in \setA$.
    \label{Rule1}
\end{itemize}

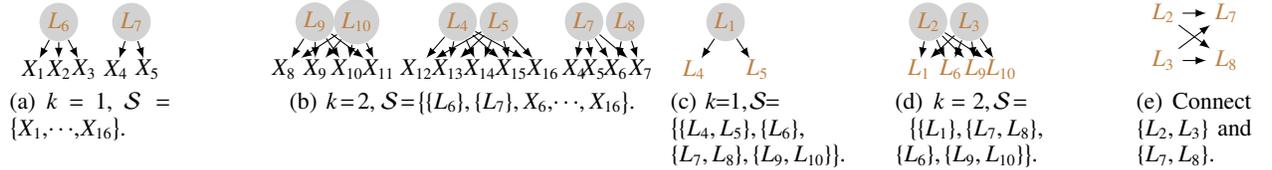
\begin{figure}[htp!]
    \centering
    \subfigure[$k\!=\!1$, $\mathcal{S} =  \big\{ X_1,\!\cdots\!,\!X_{16}\big\}$.]{
	\begin{tikzpicture}[scale=.42, line width=0.4pt, inner sep=0.1mm, shorten >=.1pt, shorten <=.1pt]
		\draw (0, 0) node[circle,fill=gray!35](L6)  {{\footnotesize\lacl\,$L_6$\,}};
		\draw (-0.8, -1.5) node(X1)  {{\footnotesize\,$X_1$\,}};
		\draw (0, -1.5) node(X2) {{\footnotesize\,$X_2$\,}};
		\draw (0.8, -1.5) node(X3) {{\footnotesize\,$X_3$\,}};
		
		\draw (2.3, 0) node[circle,fill=gray!35](L7)  {{\footnotesize\lacl\,$L_7$\,}};
		\draw (1.8, -1.5) node(X4)  {{\footnotesize\,$X_4$\,}};
		\draw (2.8, -1.5) node(X5) {{\footnotesize\,$X_5$\,}};
	   \draw[-latex] (L6) -- (X1);
	   \draw[-latex] (L6) -- (X2);
	   \draw[-latex] (L6) -- (X3);
	   \draw[-latex] (L7) -- (X4);
	   \draw[-latex] (L7) -- (X5);
	\end{tikzpicture}
    }
    \hfill
    \subfigure[$k\!=\!2$, $\mathcal{S} \!=\! \big\{ \{L_6\},\{L_7\},X_6,\!\cdots\!,X_{16}\big\}$.]{
	\begin{tikzpicture}[scale=.42, line width=0.4pt, inner sep=0.1mm, shorten >=.1pt, shorten <=.1pt]
		\draw (-0.52, 0) node[circle,fill=gray!35](L9)  {{\footnotesize\lacl\,$L_9$\,}};
		\draw (0.8, 0) node[circle,fill=gray!35](L10)  {{\footnotesize\lacl\,$L_{10}$\,}};
		\draw (-1.5, -1.5) node(X8)  {{\footnotesize\,$X_8$\,}};
		\draw (-0.5, -1.5) node(X9) {{\footnotesize\,$X_9$\,}};
		\draw (.5, -1.5) node(X10) {{\footnotesize\,$X_{10}$\,}};
		\draw (1.5, -1.5) node(X11) {{\footnotesize\,$X_{11}$\,}};
	   \draw[-latex] (L9) -- (X8);
	   \draw[-latex] (L9) -- (X9);
	   \draw[-latex] (L9) -- (X10);
	   \draw[-latex] (L9) -- (X11);
	   \draw[-latex] (L10) -- (X8);
	   \draw[-latex] (L10) -- (X9);
	   \draw[-latex] (L10) -- (X10);
	   \draw[-latex] (L10) -- (X11);
	   
	   	\draw (4, 0) node[circle,fill=gray!35](L4)  {{\footnotesize\lacl\,$L_4$\,}};
		\draw (5.3, 0) node[circle,fill=gray!35](L5)  {{\footnotesize\lacl\,$L_5$\,}};
		\draw (2.7, -1.5) node(X12)  {{\footnotesize\,$X_{12}$\,}};
		\draw (3.7, -1.5) node(X13) {{\footnotesize\,$X_{13}$\,}};
		\draw (4.7, -1.5) node(X14) {{\footnotesize\,$X_{14}$\,}};
		\draw (5.7, -1.5) node(X15) {{\footnotesize\,$X_{15}$\,}};
		\draw (6.7, -1.5) node(X16) {{\footnotesize\,$X_{16}$\,}};
	   \draw[-latex] (L4) -- (X12);
	   \draw[-latex] (L4) -- (X13);
	   \draw[-latex] (L4) -- (X14);
	   \draw[-latex] (L4) -- (X15);
	   \draw[-latex] (L4) -- (X16);
	   \draw[-latex] (L5) -- (X12);
	   \draw[-latex] (L5) -- (X13);
	   \draw[-latex] (L5) -- (X14);
	   \draw[-latex] (L5) -- (X15);
	   \draw[-latex] (L5) -- (X16);
	   
	   	\draw (8, 0) node[circle,fill=gray!35](L7)  {{\footnotesize\lacl\,$L_7$\,}};
		\draw (9.3, 0) node[circle,fill=gray!35](L8)  {{\footnotesize\lacl\,$L_8$\,}};
		\draw (7.7, -1.5) node(X4) {{\footnotesize\,$X_{4}$\,}};
		\draw (8.3, -1.5) node(X5) {{\footnotesize\,$X_{5}$\,}};
		\draw (9, -1.5) node(X6)  {{\footnotesize\,$X_6$\,}};
		\draw (9.8, -1.5) node(X7) {{\footnotesize\,$X_{7}$\,}};
		
	   \draw[-latex] (L7) -- (X4);
	   \draw[-latex] (L7) -- (X5);
	   \draw[-latex] (L8) -- (X6);
	   \draw[-latex] (L8) -- (X7);
	   \draw[-latex] (L7) -- (X6);
	   \draw[-latex] (L7) -- (X7);
	   
	\end{tikzpicture}
    }
    \subfigure[$k\!\!=\!\!1$, \!$\mathcal{S}\!\! =\!\!\!\!\! \big\{\{L_4,L_5\}, \{L_6\},$  $ \{L_7,L_8\},\{L_9,L_{10}\}\big\}$.]{
	\begin{tikzpicture}[scale=.42, line width=0.4pt, inner sep=0.1mm, shorten >=.1pt, shorten <=.1pt]
		\draw (0, 0) node[circle,fill=gray!35](L1)  {{\footnotesize\lacl\,$L_1$\,}};
		\draw (-1, -1.5) node(L4)  {{\footnotesize\lacl\,$L_4$\,}};
		\draw (1, -1.5) node(L5) {{\footnotesize\lacl\,$L_5$\,}};
	   \draw[-latex] (L1) -- (L4);
	   \draw[-latex] (L1) -- (L5);
	\end{tikzpicture}
    }
    \hfill
    ~~~
    \subfigure[$k\!\!=\!\!2$,\! \!$\mathcal{S}\! \!\!=\!\!\!~~~~ \big\{\{L_1\}, \{L_7,L_8\},$ ~~~~~~~ $\{L_6\}, \{L_9,L_{10}\}\big\}$.]{
	\begin{tikzpicture}[scale=.42, line width=0.4pt, inner sep=0.1mm, shorten >=.1pt, shorten <=.1pt]
		\draw (3.7, 0) node[circle,fill=gray!35](L2)  {{\footnotesize\lacl\,$L_2$\,}};
		\draw (5, 0) node[circle,fill=gray!35](L3)  {{\footnotesize\lacl\,$L_3$\,}};
		\draw (3.4, -1.5) node(L1) {{\footnotesize\lacl\,$L_1$\,}};
		\draw (4.4, -1.5) node(L6) {{\footnotesize\lacl\,$L_6$\,}};
		\draw (5.2, -1.5) node(L9) {{\footnotesize\lacl\,$L_9$\,}};
		\draw (6, -1.5) node(L10) {{\footnotesize\lacl\,$L_{10}$\,}};
	   \draw[-latex] (L2) -- (L1);
	   \draw[-latex] (L2) -- (L6);
	   \draw[-latex] (L2) -- (L9);
	   \draw[-latex] (L2) -- (L10);
	   \draw[-latex] (L3) -- (L1);
	   \draw[-latex] (L3) -- (L6);
	   \draw[-latex] (L3) -- (L9);
	   \draw[-latex] (L3) -- (L10);
	\end{tikzpicture}
    }
    \hfill
    ~~
    \subfigure[Connect $\{L_2,L_3\}$ and $\{L_7,L_8\}$.]{
	\begin{tikzpicture}[scale=.42, line width=0.4pt, inner sep=0.1mm, shorten >=.01pt, shorten <=.01pt]
		\draw (3.5, 0) node[circle](L2)  {{\footnotesize\lacl\,$L_2$\,}};
		\draw (5.5, 0) node[circle](L7)  {{\footnotesize\lacl\,$L_7$\,}};
		\draw (3.5, -1.5) node[circle](L3)  {{\footnotesize\lacl\,$L_3$\,}};
		\draw (5.5, -1.5) node[circle](L8)  {{\footnotesize\lacl\,$L_8$\,}};
	   \draw[-latex] (L2) -- (L7);
	   \draw[-latex] (L2) -- (L8);
	   \draw[-latex] (L3) -- (L7);
	   \draw[-latex] (L3) -- (L8);
	\end{tikzpicture}
    }
	\caption{An illustration of Algorithm \ref{findCausalClusters} that discovers new clusters in sequence (marked with gray circle) by applying \textit{findCausalClusters} to the measured variables $\setX_{\graph}$ generated from the structure in Figure \ref{fig:hierarchical-a}. Specifically, we first set $k=1$ and the active set is $\mathcal{S} = \big\{ X_1,\cdots,X_{16}\big\}$ and $\tilde{\mathcal{S}} = \mathcal{S}$, and we can find the clusters in (a), and no further cluster can be found with $k=1$. Then we increase $k$ to $2$ with the active set $\mathcal{S} = \{\big\{L_6\},\{L_7\},X_6,\cdots,X_{16}\big\}$ and $\tilde{\mathcal{S}} = \mathcal{S}$, and then we can find the clusters in (b).  Then, the active set is $\mathcal{S} = \big\{\{L_4, L_5\},\{L_6\},\{L_7,L_8\},\{L_9,L_{10}\}\big\}$ and we set back $k=1$, and when $\tilde{\mathcal{S}}=\{\{L_4,L_5\}, X_1,\cdots,X_{11}\}$ we find the cluster in (c).  Note that when testing the rank over $\{L_4,L_5\}$ against other variables, we use their measured pure descendants in the currently estimated graph instead. The above procedure is repeated to further find the cluster in (d). Finally, when there are no enough variables for testing, we connect the elements in the active variable set: connecting $\{L_2,L_3\}$ to $\{L_7,L_8\}$ in (e).}
	\label{Fig:findCausalClusters_exp}
\end{figure}

Later in Phase II, we will show that Rule 1 may not correctly identify the latent atomic cover in some cases, and, accordingly, we will further provide an efficient revision procedure in Phase II. At the current phase, we use Rule 1, together with certain search procedures, to identify the clusters and latent atomic covers. The detailed search procedure of Phase I is given in Algorithm \ref{findCausalClusters} (\textit{findCausalClusters}), which tests for rank deficiency recursively to discover clusters of variables and their latent atomic clusters. The set of active variables $\mathcal{S}$ is set to $\measures$ initially (line 1) and will be updated as the search goes on (line 17). We start to identify the latent atomic cover with size $k=1$ (line 2). We consider any subset of the latent atomic covers in $\mathcal{S}$ and replace them with their pure children, resulting in $\tilde{\mathcal{S}}$ (lines 5-6). 
Then we draw a subset of variables $\setA \subset \tilde{\mathcal{S}}$ with cardinality at least $k+1$ and conduct a rank deficiency test of $\setA$ against $\mathcal{S} \backslash \setA$ by estimating the rank of $\texttt{rank}(\Sigma_{\setX_A,\setX_B})$ (lines 8-9). Note that here $\setX_A$ and $\setX_B$ are the measured pure descendants of $\setA$ and $\setB$, respectively, in the currently learned graph. This step is repeated until all subsets are tested (lines 7-11). If rank deficiency is found, we merge the overlapping groups into a cluster and add latent covers over the cluster (lines 12-17). We further reset $k=1$ and resume the search (lines 18-19). Otherwise, if no latent cover is found, we increment $k = k+1$ (line 21). This procedure is repeated until no more clusters are found. Finally, we connect the elements in $\mathcal{S}$ into a chain structure (line 23). Figure \ref{Fig:findCausalClusters_exp} illustrates an example procedure of finding new clusters by applying \textit{findCausalClusters} to the measured variables $\setX_{\graph}$ generated from the structure in Figure \ref{fig:hierarchical-a}.

\begin{algorithm}[!htp]
\setlength{\textfloatsep}{-0pt}
\SetKwInOut{Input}{Input}
\SetKwInOut{Output}{Output}
\SetKwFunction{findCausalClusters}{\texttt{findCausalClusters}}

\Input{Data from a set of measured variables $\measures$}
\Output{Graph $\graph'$}

Active set $\mathcal{S} \gets \measures$\;
$k \gets 1$\;

\Repeat{no more clusters are found}{
      \Repeat{all subsets $\tilde{\mathbf{L}}$ exhausted}{
       draw a set of latent atomic covers $\tilde{\mathbf{L}} \subset \mathcal{S}$\;
       let $\tilde{\mathcal{S}} = (\mathcal{S} \backslash \tilde{\mathbf{L}}) \cup (\cup_{L_i \in \tilde{\mathbf{L}}} PCh(L_i) )$\;
       
        \Repeat{all subsets $\mathbf{A}$ exhausted}{
        draw a set of test variables $\mathbf{A} \subset \tilde{\mathcal{S}}$, with $\left\Vert \mathbf{A} \right\Vert \geq k+1$, and let $\mathbf{B} \leftarrow \tilde{\mathcal{S}} \backslash \mathbf{A}$\;
        
        $k' \leftarrow \texttt{rank}(\Sigma_{\setX_A, \setX_B})$ estimated by rank deficiency test\;
        
            \lIf{$k' < k + 1$}{
                rank deficiency found and keep track of this set $\setA$
            }
        }
        
        \If{any groups with rank deficiency are found}{
            merge overlapping clusters\;
            identify lowest rank $k'$ found\;
            
            \ForEach{discovered cluster of variables $\setA$ with rank $k'$}{
                create latent cover $\setL$ with cardinality $k'$ as parents of $\setA$\;
                $\mathcal{S} = (\mathcal{S} \backslash \setA) \cup \setL$\;
            }
            $k=1$\;
            $\textbf{break}$\;
        }
    }
    \lIf{no group with rank deficiency is found}{
    $k = k+1$}
 }
for all $S_i \in \mathcal{S}$, connect them into a chain structure\;

\Return{$\graphp$}
\caption{Phase I: findCausalClusters}
\label{findCausalClusters}
\end{algorithm}

\subsection{Phase II: Refining Clusters}
As we have mentioned above, the naive assignment of causal clusters in Algorithm \ref{findCausalClusters} (\textit{findCausalClusters}) may not be correct in some cases. In this section, we provide a precise characterization of the cases where \textit{findCausalClusters} incorrectly clusters variables and, accordingly, propose an efficient algorithm to correct such cases.

\begin{figure}[htp!]
\setlength{\abovecaptionskip}{-3pt}
\setlength{\belowcaptionskip}{-5pt}
    \centering
    \subfigure[Original graph]{
        \begin{tikzpicture}[scale=.55, line width=0.4pt, inner sep=0.2mm, shorten >=.1pt, shorten <=.1pt]
		\draw (0, 2) node(L1)  {{\footnotesize\lacl\,$L_1$\,}};
		\draw (1.5, 2) node(L2)  {{\footnotesize\lacl\,$L_2$\,}};
		\draw (-1.5, 2) node(L3) {{\footnotesize\lacl\,$L_3$\,}};
		\draw (3, 1.5) node(L4)  {{\footnotesize\lacl\,$L_4$\,}};
		\draw (3, 2.5) node(L5)  {{\footnotesize\lacl\,$L_5$\,}};
		\draw (-3, 1.5) node(L6) {{\footnotesize\lacl\,$L_6$\,}};
		\draw (-3, 2.5) node(L7) {{\footnotesize\lacl\,$L_7$\,}};
		\draw (5, 1) node(X1) {{\footnotesize\,$X_1$\,}};
		\draw (5, 1.7) node(X2) {{\footnotesize\,$X_2$\,}};
		\draw (5, 2.4) node(X3) {{\footnotesize\,$X_3$\,}};
		\draw (5, 3) node(X4) {{\footnotesize\,$X_4$\,}};
		\draw (-5, 1) node(X5) {{\footnotesize\,$X_5$\,}};
		\draw (-5, 1.7) node(X6) {{\footnotesize\,$X_6$\,}};
		\draw (-5, 2.4) node(X7) {{\footnotesize\,$X_7$\,}};
		\draw (-5, 3) node(X8) {{\footnotesize\,$X_8$\,}};
		\draw (1.5, 1) node(X9) {{\footnotesize\,$X_9$\,}};
		\draw (0, 1) node(X10) {{\footnotesize\,$X_{10}$\,}};
		\draw (-1.5, 1) node(X11) {{\footnotesize\,$X_{11}$\,}};
		
		\draw[-latex] (L1) -- (L2);
		\draw[-latex] (L1) -- (L3);
		\draw[-latex] (L2) -- (L4);
		\draw[-latex] (L2) -- (L5);
		\draw[-latex] (L3) -- (L6);
		\draw[-latex] (L3) -- (L7);
		\draw[-latex] (L2) -- (X9);
		\draw[-latex] (L1) -- (X10);
		\draw[-latex] (L3) -- (X11);
		\draw[-latex] (L4) -- (X1);
		\draw[-latex] (L4) -- (X2);
		\draw[-latex] (L4) -- (X3);
		\draw[-latex] (L4) -- (X4);
		\draw[-latex] (L5) -- (X1);
		\draw[-latex] (L5) -- (X2);
		\draw[-latex] (L5) -- (X3);
		\draw[-latex] (L5) -- (X4);
		\draw[-latex] (L6) -- (X5);
		\draw[-latex] (L6) -- (X6);
		\draw[-latex] (L6) -- (X7);
		\draw[-latex] (L6) -- (X8);
		\draw[-latex] (L7) -- (X5);
		\draw[-latex] (L7) -- (X6);
		\draw[-latex] (L7) -- (X7);
		\draw[-latex] (L7) -- (X8);
	\end{tikzpicture}
	\label{fig:refineClusters-a}
    }
    ~~
\subfigure[Output from findCausalClusters]{
    \begin{tikzpicture}[scale=.55, line width=0.4pt, inner sep=0.2mm, shorten >=.1pt, shorten <=.1pt]
		\draw (0.5, 3.5) node(L1)  {{\footnotesize\lacl\,$L_1'$\,}};
		\draw (-0.5, 3.5) node(L2) {{\footnotesize\lacl\,$L_2'$\,}};
		\draw (3.5, 1.5) node(L3)  {{\footnotesize\lacl\,$L_3'$\,}};
		\draw (3.5, 2.5) node(L4)  {{\footnotesize\lacl\,$L_4'$\,}};
		\draw (-3.5, 1.5) node(L5) {{\footnotesize\lacl\,$L_5'$\,}};
		\draw (-3.5, 2.5) node(L6) {{\footnotesize\lacl\,$L_6'$\,}};
		\draw (-2, 2) node(L7) {{\footnotesize\lacl\,$L_7'$\,}};
		\draw (2, 2) node(L8) {{\footnotesize\lacl\,$L_8'$\,}};
		
		\draw (5, 1) node(X1) {{\footnotesize\,$X_1$\,}};
		\draw (5, 1.7) node(X2) {{\footnotesize\,$X_2$\,}};
		\draw (5, 2.4) node(X3) {{\footnotesize\,$X_3$\,}};
		\draw (5, 3) node(X4) {{\footnotesize\,$X_4$\,}};
		\draw (-5, 1) node(X5) {{\footnotesize\,$X_5$\,}};
		\draw (-5, 1.7) node(X6) {{\footnotesize\,$X_6$\,}};
		\draw (-5, 2.4) node(X7) {{\footnotesize\,$X_7$\,}};
		\draw (-5, 3) node(X8) {{\footnotesize\,$X_8$\,}};
		\draw (1.5, 1) node(X9) {{\footnotesize\,$X_9$\,}};
		\draw (0, 1) node(X10) {{\footnotesize\,$X_{10}$\,}};
		\draw (-1.5, 1) node(X11) {{\footnotesize\,$X_{11}$\,}};
		
		\draw[-latex] (L1) -- (X9);
		\draw[-latex] (L1) -- (X10);
		\draw[-latex] (L1) -- (X11);
		\draw[-latex] (L2) -- (X9);
		\draw[-latex] (L2) -- (X10);
		\draw[-latex] (L2) -- (X11);
		
		\draw[-latex] (L3) -- (X1);
		\draw[-latex] (L3) -- (X2);
		\draw[-latex] (L3) -- (X3);
		\draw[-latex] (L3) -- (X4);
		\draw[-latex] (L4) -- (X1);
		\draw[-latex] (L4) -- (X2);
		\draw[-latex] (L4) -- (X3);
		\draw[-latex] (L4) -- (X4);

		\draw[-latex] (L5) -- (X5);
		\draw[-latex] (L5) -- (X6);
		\draw[-latex] (L5) -- (X7);
		\draw[-latex] (L5) -- (X8);
		\draw[-latex] (L6) -- (X5);
		\draw[-latex] (L6) -- (X6);
		\draw[-latex] (L6) -- (X7);
		\draw[-latex] (L6) -- (X8);

		\draw[-latex] (L7) -- (L5);
		\draw[-latex] (L7) -- (L6);
		\draw[-latex] (L8) -- (L3);
		\draw[-latex] (L8) -- (L4);
		
		\draw[-latex] (L1) -- (L7);
		\draw[-latex] (L2) -- (L7);
		\draw[-latex] (L1) -- (L8);
		\draw[-latex] (L2) -- (L8);
	\end{tikzpicture}
	\label{fig:refineClusters-b}
    }
	\caption{An example where findCausalClusters fails.}
	\label{fig:refineClusters}
\end{figure}
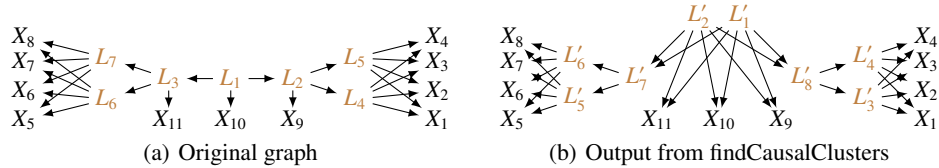

Take Figure \ref{fig:refineClusters-a} as an example to illustrate the issue. By applying \textit{findCausalClusters} to the measured variables $\setX_{\graph}$, at $k=2$, we discovered that $\{X_9, X_{10}, X_{11} \}$ form a cluster and then set a 2-latent atomic cover $\{L_1',L_2'\}$ over it (Figure \ref{fig:refineClusters-b}). This is not correct, because $X_9, X_{10}, X_{11}$ actually belong to three different 1-latent atomic covers, $L_1, L_2, L_3$, respectively. The incorrect clustering and covering are because when discovering the rank-deficiency set $\{X_9, X_{10}, X_{11} \}$, the latent covers $\{L_4, L_5\}, \{L_6, L_7\}$ have not been identified yet. If $\{L_4, L_5\}$ had already been discovered, we would correctly find that $\{X_9, L_4, L_5\}$ form a 1-latent atomic cover, and the algorithm would proceed correctly thereafter. More generally, \textit{findCausalClusters} may set incorrect cluster for the rank-deficiency set $\{X_9, X_{10}, X_{11} \}$, in the case when their parents $L_1, L_2, L_3$ split the graph into two or more disconnected graphs. This result is formally stated in the following definition and theorem.

\begin{definition}[Bond Set]
Consider a set of measured variables $\setX \subseteq \measures$, and a minimal set of latent variables $\mathbf{L} \subseteq \latents$ that d-separate $\setX$ from all other measures $\setX' \coloneqq \measures \backslash \setX$. We say that $\setX$ is a bond set if $\mathbf{L}$ also d-separates some partition $\setX_A \subset \setX', \setX_B \subset \setX'$ from one another.
\end{definition}

From the previous example, we have seen that with the existence of bond sets, \textit{findCausalClusters} may end up with incorrect clusters and covers. The following theorem shows that the presence of bond sets is the only reason for incorrect clusters and covers with \textit{findCausalClusters}.

\begin{theorem}[Correct Cluster Condition]
Suppose $\graph$ is an IL$^2$H graph with measured variables $\measures$. Consider the output $\graphp$ from applying \textit{findCausalClusters} over $\measures$. If none of the clusters in $\graphp$ is the bond set in $\graph$, then all latent atomic covers have been correctly identified. 
\label{The:correct_cluster}
\end{theorem}

However, without access to the true graph $\graph$, we are not able to identify which clusters formed from \textit{findCausalClusters} are bond sets. Fortunately, the following properties of the clusters formed over bond sets provide a way to remove bond sets in our discovered graph even without such knowledge, so that the clusters and latent covers can be correctly identified.


\begin{theorem}[Correcting Clusters]
Denote by $\graphp$ the output from \textit{findCausalClusters} and by $\graph$ the true graph. For a latent atomic cover $\setL'$ in $\graphp$, if the measured pure descendants of $\setL'$ is a bond set in the true graph $\graph$, then there exist a set of siblings $\setS$ of $\setL'$ in $\graphp$, a set of children $\setC$ of $\setL'$, and a set of grandparents $\setP$ of $\setL'$, such that $\measuredp(\setS \cup \setC \cup \setP)$ forms a cluster that is not a bond set in $\graph$.
\label{The:remove_bond}
\end{theorem}

The above theorem shows that whenever forming a bond set, the siblings and grandparents of the latent cover are the key to correcting the incorrect clusters and covers. Specifically, we will remove such a cover and use its children to form new clusters with the siblings and grandparents (see details in Algorithm \ref{refineClusters}), and Theorem \ref{The:remove_bond} guarantees that these new clusters will not contain bond sets, and thus, providing  correct clusters.

\begin{example}
In Figure \ref{fig:refineClusters-b}, the formed cluster $\{X_9,X_{10},X_{11}\}$ is the bond set in the true graph in Figure \ref{fig:refineClusters-a}, and the covers $\{L'_1,L'_2\}$, $\{L'_7\}$, and $\{L'_8\}$ are not correct. Illustration of the algorithm for this example, including refining the incorrect clusters and covers, is given in Appendix A.16. 
\end{example}

Furthermore, with Theorem \ref{The:remove_bond}, we give the following rule for correcting the clusters and covers.
\begin{itemize}[leftmargin=30pt,align=left,itemsep=0pt,topsep=0pt]
    \item[\textbf{Rule 2:}] For each discovered latent atomic cover $\setL$, let $\mathbf{V} = \grandparentsp(\setL) \bigcup \siblingsp(\setL) \bigcup \childrenp(\setL)$ and apply \textit{findCausalClusters} to $\mathbf{V}$ to refine the clusters.
    \label{Rule2}
\end{itemize}

Before introducing the detailed refining algorithm \textit{refineClusters} based on Rule 2, we first introduce an operator \textit{makeRoot} that will be used in the algorithm, which will not change the rank deficiency constraints (Lemma \ref{Lem:rank invar}).

\begin{definition}[\textit{makeRoot}]
Give a graph $\graphp$ and a latent atomic cover $\setL$ in $\graphp$. A makeRoot operator of $\setL$, denoted by $makeRoot(\setL)$, reorients all outgoing edges of $\setL$ to $\setL$, such that $\setL$ is a root variable.
\end{definition}

\begin{lemma}[Rank Invariance]
Denote by $\graphp$ the output from \textit{findCausalClusters} and by $\setL$ a latent atomic cover in $\graphp$. Then all rank constraints, that are possible to be executed by \textit{findCausalClusters} prescribed by $\graphp$, before and after the operator $makeRoot(\setL)$ are identical. 
\label{Lem:rank invar}
\end{lemma}

Algorithm \ref{refineClusters} introduces the procedure of refining clusters based on Rule 2. Starting with the output $\graphp$ from \textit{findCausalClusters}, we proceed in a breadth-first search from the root variable (lines 1-4, 6). Because the root is trivially not over a bond set, Theorem \ref{The:remove_bond} guarantees that for each child of the root, we will form new clusters that are not over bond sets. For each cover $\setL$, if it has only one child, we will recursively consider the children of its child instead, since if $\setL$ is over a bond set then its single child will also be. We add the first set of children into the search set $\setV$ (line 5). We construct the search set (line 7) and remove the covers we are refining (lines 8-9). Finally, we make $\setL$ the root (lines 9-10) and conduct the search to form new clusters (lines 11-12). The algorithm ends after refining every latent cover in $\graphp$. With this refining procedure, we will derive the correct latent covers for the graph in Figure \ref{fig:refineClusters-a}.

\begin{algorithm}[!htp]
\SetAlgoLined
\SetKwInOut{Input}{Input}
\SetKwInOut{Output}{Output}
\SetKwFunction{refineClusters}{\texttt{refineClusters}}
\Input{Output graph $\graph'$ from Phase I}
\Output{Refined graph $\graph'$}

\SetInd{0.5em}{0.5em}
\Indm\nonl\refineClusters($\graphp$):\\
\Indp
\SetInd{0.5em}{1.0em}
Let $Q$ be an empty queue\;
$Q$.enqueue($\purechildrenp(Root(\mathcal{G'}))$)\;

\Repeat{$Q$ is empty}{
    $\setL \leftarrow$ $Q$.dequeue()\;
    
    \lWhile{$\childrenp(\setL)$ is a single atomic cover}{
        remove $\childrenp(\setL)$ and add $\childrenp(\childrenp(\setL))$ as children of $\setL$
    }
    \lFor{each atomic cover $C \in \childrenp(\setL)$}{
        $Q$.enqueue($\setC$)
    }
    $\mathbf{V} \leftarrow \grandparentsp(\setL) \bigcup \siblingsp(\setL) \bigcup \childrenp(\setL)$ \tcp*{\footnotesize variables for finding new clusters}
    $\mathbf{R} \leftarrow \setL \bigcup \parents(\setL)$ \tcp*{\footnotesize atomic covers to be removed}
    $\graphp \gets \texttt{makeRoot}_{\graphp}(\setL)$\;
    remove $\mathbf{R}$ and all adjoining edges from $\mathcal{G'}$\;
    
    $\mathcal{G''} \leftarrow \texttt{findCausalClusters}(\setV)$
    
    update $\graphp$ with new clusters from $\mathcal{G''}$ \tcp*{\footnotesize find new clusters} 
}
\Return{$\graphp$}
\caption{Phase II: refineClusters}
\label{refineClusters}
\end{algorithm}

\subsection{Phase III: Refining Edges}

With \textit{findCausalClusters} and \textit{refineClusters}, the output $\graphp$ correctly identifies the latent variables.  Moreover, $\graphp$ correctly identifies the following d-separation: for every $\setL \in \setL_{\graphp}$, its parents $\parentsp(\setL)$ d-separates $\setL$ and its descendants from the ancestors of $\parentsp(\setL)$, and thus, there cannot be any edges from each $\setL$ to any of its ancestors beyond its own parents.  
However, $\graphp$ may still have incorrect edges \textit{locally}. Specifically, each time we create an atomic cover $\setL$ in \textit{findCausalClusters}, the implicit assumption is that each of its children is conditionally independent of the other children given the parents $\setL$. However, this assumption is not necessarily true, as (i) the children of $\setL$ may be directly connected to one another, and (ii) $\setL$ may only be directly connected to a subset of its children. In other words, previous steps did not consider condition independence relationships across clusters. Hence, we need a further step to correct edges over each $\setL$ and its children.

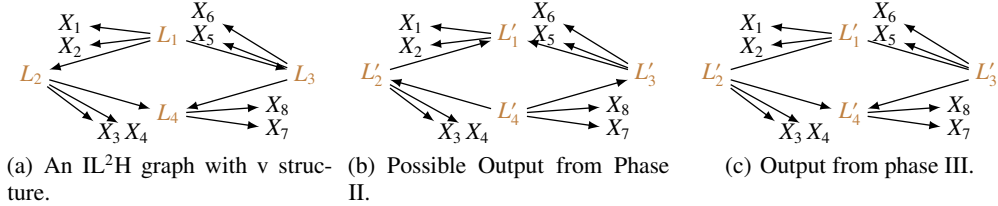
\begin{figure}[htp!]
    \centering
    \subfigure[An IL$^2$H graph with v structure.]{
    \begin{tikzpicture}[scale=.52, line width=0.5pt, inner sep=0.2mm, shorten >=.1pt, shorten <=.1pt]
		\draw (0, 3) node(L1)  {{\footnotesize\lacl\,$L_1$\,}};
		\draw (-3.5, 2) node(L2) {{\footnotesize\lacl\,$L_2$\,}};
		\draw (3.5, 2) node(L3) {{\footnotesize\lacl\,$L_3$\,}};
		\draw (0, 1) node(L4) {{\footnotesize\lacl\,$L_4$\,}};
		
		\draw (-2.5, 3.3) node(X1)  {{\footnotesize\,$X_1$\,}};
		\draw (-2.5, 2.7) node(X2)  {{\footnotesize\,$X_2$\,}};
		\draw (-1.5, 0.5) node(X3) {{\footnotesize\,$X_3$\,}};
		\draw (-0.8, 0.5) node(X4) {{\footnotesize\,$X_4$\,}};
		\draw (0.9, 3) node(X5)  {{\footnotesize\,$X_5$\,}};
		\draw (0.9, 3.6) node(X6)  {{\footnotesize\,$X_6$\,}};
		\draw (2.8, 0.6) node(X7)  {{\footnotesize\,$X_7$\,}};
		\draw (2.8, 1.2) node(X8)  {{\footnotesize\,$X_8$\,}};
		\draw[-latex] (L1) -- (L2);
		\draw[-latex] (L2) -- (L4);
		\draw[-latex] (L1) -- (L3);
		\draw[-latex] (L3) -- (L4);
		\draw[-latex] (L1) -- (X1);
		\draw[-latex] (L1) -- (X2);
		\draw[-latex] (L2) -- (X3);
		\draw[-latex] (L2) -- (X4);
		\draw[-latex] (L3) -- (X5);
		\draw[-latex] (L3) -- (X6);
		\draw[-latex] (L4) -- (X7);
		\draw[-latex] (L4) -- (X8);
	\end{tikzpicture}
	\label{fig:refineEdges-a}
    }
    ~
    \subfigure[Possible Output from Phase II.]{
    \begin{tikzpicture}[scale=.52, line width=0.5pt, inner sep=0.2mm, shorten >=.1pt, shorten <=.1pt]
		\draw (0, 3) node(L1)  {{\footnotesize\lacl\,$L_1'$\,}};
		\draw (-3.5, 2) node(L2) {{\footnotesize\lacl\,$L_2'$\,}};
		\draw (3.5, 2) node(L3) {{\footnotesize\lacl\,$L_3'$\,}};
		\draw (0, 1) node(L4) {{\footnotesize\lacl\,$L_4'$\,}};
		
		\draw (-2.5, 3.3) node(X1)  {{\footnotesize\,$X_1$\,}};
		\draw (-2.5, 2.7) node(X2)  {{\footnotesize\,$X_2$\,}};
		\draw (-1.5, 0.5) node(X3) {{\footnotesize\,$X_3$\,}};
		\draw (-0.8, 0.5) node(X4) {{\footnotesize\,$X_4$\,}};
		\draw (0.9, 3) node(X5)  {{\footnotesize\,$X_5$\,}};
		\draw (0.9, 3.6) node(X6)  {{\footnotesize\,$X_6$\,}};
		\draw (2.8, 0.6) node(X7)  {{\footnotesize\,$X_7$\,}};
		\draw (2.8, 1.2) node(X8)  {{\footnotesize\,$X_8$\,}};
		\draw[-latex] (L2) -- (L1);
		\draw[-latex] (L3) -- (L1);
		\draw[-latex] (L4) -- (L3);
		\draw[-latex] (L4) -- (L2);
		\draw[-latex] (L1) -- (X1);
		\draw[-latex] (L1) -- (X2);
		\draw[-latex] (L2) -- (X3);
		\draw[-latex] (L2) -- (X4);
		\draw[-latex] (L3) -- (X5);
		\draw[-latex] (L3) -- (X6);
		\draw[-latex] (L4) -- (X7);
		\draw[-latex] (L4) -- (X8);
	\end{tikzpicture}
	\label{fig:refineEdges-b}
    }
    ~
    \subfigure[Output from phase III.]{
    \begin{tikzpicture}[scale=.52, line width=0.5pt, inner sep=0.2mm, shorten >=.1pt, shorten <=.1pt]
		\draw (0, 3) node(L1)  {{\footnotesize\lacl\,$L_1'$\,}};
		\draw (-3.5, 2) node(L2) {{\footnotesize\lacl\,$L_2'$\,}};
		\draw (3.5, 2) node(L3) {{\footnotesize\lacl\,$L_3'$\,}};
		\draw (0, 1) node(L4) {{\footnotesize\lacl\,$L_4'$\,}};
		
		\draw (-2.5, 3.3) node(X1)  {{\footnotesize\,$X_1$\,}};
		\draw (-2.5, 2.7) node(X2)  {{\footnotesize\,$X_2$\,}};
		\draw (-1.5, 0.5) node(X3) {{\footnotesize\,$X_3$\,}};
		\draw (-0.8, 0.5) node(X4) {{\footnotesize\,$X_4$\,}};
		\draw (0.9, 3) node(X5)  {{\footnotesize\,$X_5$\,}};
		\draw (0.9, 3.6) node(X6)  {{\footnotesize\,$X_6$\,}};
		\draw (2.8, 0.6) node(X7)  {{\footnotesize\,$X_7$\,}};
		\draw (2.8, 1.2) node(X8)  {{\footnotesize\,$X_8$\,}};
		\draw (L1) -- (L2);
		\draw[-latex] (L2) -- (L4);
		\draw (L1) -- (L3);
		\draw[-latex] (L3) -- (L4);
		\draw[-latex] (L1) -- (X1);
		\draw[-latex] (L1) -- (X2);
		\draw[-latex] (L2) -- (X3);
		\draw[-latex] (L2) -- (X4);
		\draw[-latex] (L3) -- (X5);
		\draw[-latex] (L3) -- (X6);
		\draw[-latex] (L4) -- (X7);
		\draw[-latex] (L4) -- (X8);
	\end{tikzpicture}
	\label{fig:refineEdges-c}
    }
    \caption{Illustrations for refineEdges.}
    \label{fig:refineEdges}
\end{figure}

For example, consider the true graph in Figure \ref{fig:refineEdges-a}, which ends up with the graph in Figure \ref{fig:refineEdges-b} with the first two phases. However, note that the first two phases did not consider the d-separation between $L'_2$ and $L'_3$, and thus the edges among $L_1', L_2', L_3', L_4'$ may not be correct, including the v structure. In particular, in the true graph, $L_1$ d-separates $L_2$ from $L_3$, while $\{L_1, L_4\}$ does not d-separate $L_2$ from $L_3$, but these d-separations are not reflected in the discovered $\graphp$ in the first two phases. 
Thus, we need to refine the edges. To this end, we first set $L_1', L_2', L_3'$ and $L_2', L_3', L_4'$ to be fully connected, and consider testing $\mathcal{A} = \{L_2', X_1\}$ against $\mathcal{B} = \{L_3', X_2\}$, where we partition the children of $L_1'$ into two sets and put them in $\mathcal{A}$ and $\mathcal{B}$, respectively. By doing so, we force $L_1'$ in $\graph'$ to be in the separating set. Since $\texttt{rank}(\Sigma_{\mathcal{A},\mathcal{B}}) = 1$, it implies that no other variable is in the separating set, and therefore we can conclude that $L_1'$ d-separates $L_2'$ from $L_3'$. This principle is characterized in the following lemma. 

\begin{lemma}[Cross-Cover Test]
Given a set of variables $\mathcal{S}$, consider two latent atomic covers $\setL_A, \setL_B \in \mathcal{S}$, and a potential separating set $\setC = \{ \setL_{C_i} \} \subseteq \mathcal{S} \backslash \{\setL_A, \setL_B\}$. For each $\setL_{C_i}$, consider $\setC_i^A, \setC_i^B \subseteq PCh(\setL_{C_i})$ with $\setC_i^A, \setC_i^B \neq \emptyset$ and $\setC_i^A \cap \setC_i^B = \emptyset$, and denote the cardinality $k_i^A \coloneqq min(|\setL_{C_i}|, |\setC_i^A|)$, $k_i^B \coloneqq min(|\setL_{C_i}|, |\setC_i^B|)$, respectively. Then there is no edge between $\setL_A$ and $\setL_B$ if and only if there exists a separating set $\setC$ such that $\texttt{rank}(\Sigma_{\mathcal{A}, \mathcal{B}}) < min(|\setL_A| + \sum_i k_i^A, |\setL_B| + \sum_i k_i^B)$, where $\mathcal{A} = \{\setL_A, \setC_1^A, \setC_2^A, ...\}$ and $\mathcal{B} = \{\setL_B, \setC_1^B, \setC_2^B, ...\}$. In this case, we say that $\setC$ satisfies the cross-cover test of $\setL_A$ against $\setL_B$. 
\label{Crosscover}
\end{lemma}

Note that in order to find rank deficiency when performing the cross-cover test, $\setL_{C_i}$ needs to satisfy the third condition in Condition \ref{Condition:R2H}. 
Based on Lemma \ref{Crosscover}, we use the following rule to refine the edges.
\begin{itemize}[leftmargin=30pt,align=left,itemsep=0pt,topsep=0pt]
    \item[\textbf{Rule 3:}] For a pair of latent covers $(\setL_A, \setL_B)$, let $\mathcal{A} \gets \{ \setL_A, \setC_1^A, \setC_2^A, ...\}$ and $\mathcal{B} \gets \{ \setL_B, \setC_1^B, \setC_2^B, ...\}$. If there exists such $\mathcal{A}, \mathcal{B}$ such that $\texttt{rank}_{\graphp}(\Sigma_{\mathcal{A}, \mathcal{B}})$ is rank deficient, then remove all edges between $\setL_A, \setL_B$ in $\graph'$.
    \label{Rule4}
\end{itemize}

Furthermore, with Rule 3, Algorithm \ref{conditionalIndependenceTest} (\textit{CrossCoverTest}) gives the procedure of refining the edges over a set of latent variables $\mathcal{S}$ to correct the causal skeleton. It first fully connects the latent covers in $\mathcal{S}$ (line 2). Then for every pair of latent covers, it performs the cross-cover test (lines 3-17). If rank deficiency is found, then remove the corresponding edges (lines 11-12).

Furthermore, we are going to identify the v structures among latent atomic covers. In the output $\graphp$ from phase II, for any $\setL'$ with a child $\setC'_i$ and parent $\setP'$, it is not possible to have a collider $\setC'_i \rightarrow \setL' \leftarrow \setP'$ in the ground-truth graph, because in this case the cluster would not have been rank deficient. Therefore, the only v structures in the graph are amongst the variables in $\setL' \cup \purechildrenp(\setL')$, and similar to \textit{crossCoverTest}, we only need to test for v structures locally. Continue to consider the example in Figure \ref{fig:refineEdges}. The edge between $L_2'$ and $L_3'$ is missing because the rank over $\{L_2', X_1\}$ and $\{L_3', X_2\}$ is $1$, implying that $L_1'$ d-separates $L_2'$ from $L_3'$, as is done in \textit{crossCoverTest}. Now, since $L_2'-L_4'-L_3'$ forms an unshielded triplet, we want to test if a collider exists at $L_4'$. We find that the rank over $\mathcal{A}=\{L_2', L_4', X_1\}$ and $\mathcal{B}=\{L_3', X_2\}$ is $2 > 1$, and the rank over $\mathcal{A}=\{L_2', X_1\}$ and $\mathcal{B}=\{L_3', L_4', X_2\}$ is $2 > 1$, so $L'_2 \rightarrow L'_4 \leftarrow L'_3$. For general cases, the rule for finding v structures are formulated in the following lemma and Algorithm \ref{Alg:findColliders} (\textit{findColliders}). Specifically, in Algorithm \ref{Alg:findColliders} (\textit{findColliders}), for every unshielded triangle $\setL_1 - \setL_3 - \setL_2$, it performs v structure tests and compares with the rank which does not involve $\setL_3$ (lines 2-5). If the rank that involves $\setL_3$ is larger, then the unshielded triangle forms a v structure (lines 6-7).

\begin{lemma}[V-Structure Test]
For any unshielded triangle $\setL_A-\setL_C-\setL_B$, let $\mathcal{A}, \mathcal{B}$ be the set of variables in Lemma \ref{Crosscover} such that $\Sigma_{\mathcal{A}, \mathcal{B}}$ was rank deficient.
Let $k = \texttt{rank}(\Sigma_{\mathcal{A}, \mathcal{B}})$, $k_1 = \texttt{rank}(\Sigma_{\mathcal{A}\cup \setL_C, \mathcal{B}})$, and $k_2 = \texttt{rank}(\Sigma_{\mathcal{A}, \mathcal{B}\cup \setL_C})$. Then, $\setL_A \rightarrow \setL_B \leftarrow \setL_C$ if and only if $k<\min(k_1, k_2)$.
\label{Lem:v-str}
\end{lemma}

\begin{algorithm}[htp!]
\label{conditionalIndependenceTest}
\SetKwInOut{Input}{Input}
\SetKwInOut{Output}{Output}
\SetKwFunction{conditionalIndependenceTest}{\texttt{crossCoverTest}}
\Input{A set of latent variables $\mathcal{S}$, currently learned graph $\graphp$}
\Output{An edgeset $\mathcal{E}$ among the variables}

\SetInd{0.5em}{0.5em}
\Indm\nonl\conditionalIndependenceTest($\mathcal{S}$, $\graphp$):\\
\Indp
\SetInd{0.5em}{1.0em}
$\mathcal{E} \gets \emptyset$\;

add undirected edges between every pair of latent atomic covers in $\mathcal{S}$ to $\mathcal{E}$\;

\ForEach{
    pair of latent atomic covers $\setL_A,\setL_B \in \mathcal{S}$
}{
    k = 0\;
    
    \Repeat{
        $k > $ number of variables in $\mathcal{S} \backslash \{ \setL_A, \setL_B \}$
    }{
        \Repeat{
            all sets $\setC$ with $k$ atomic covers tested
        }{
            draw a potential separating set of $k$ atomic covers $\setC = \{ \setC_1, \setC_2, \cdots, \setC_k \} \subseteq \mathcal{S} \backslash \{\setL_A, \setL_B\}$\;
            
            \ForEach{
                atomic cover $\setC_i \in \setC$
            }{
                partition $\purechildrenp(\setC_i)$ into $\setC_i^A, \setC_i^B$ \tcp*{Remark on clever choice}
            }
            $\mathcal{A} \gets \{ \setL_A, \setC_1^A, \setC_2^A, ...\}$ and $\mathcal{B} \gets \{ \setL_B, \setC_1^B, \setC_2^B, ...\}$\;
            
            \If{
                there exists such $\mathcal{A}, \mathcal{B}$ such that $\texttt{rank}_{\graphp}(\Sigma_{\mathcal{A}, \mathcal{B}})$ is rank deficient
            }{
                remove all edges between $\setL_A, \setL_B$ in $\mathcal{E}$\;
                
                break\;
            }
        }
        \If{rank deficiency found}
        {break\;}
        $k \mathrel{+}= 1$\;
    }
}
\Return{Edgeset $\mathcal{E}$}
\caption{crossCoverTest}
\label{conditionalIndependenceTest}
\end{algorithm}

\begin{algorithm}[htp!]
\label{findColliders}
\SetKwInOut{Input}{Input}
\SetKwInOut{Output}{Output}
\SetKwFunction{findColliders}{\texttt{findColliders}}
\Input{A set of latent variables $\mathcal{S}$, edgeset $\mathcal{E}$, currently learned graph $\graphp$}
\Output{A set of colliders $\mathcal{C}$}

\SetInd{0.5em}{0.5em}
\Indm\nonl\findColliders($\mathcal{S}$, $\mathcal{E}$, $\graphp$):\\
\Indp
\SetInd{0.5em}{1.0em}
Collider set $\mathcal{C} \gets \emptyset$\;

\ForEach{unshielded triangle $\setL_1-\setL_3-\setL_2$ in $\mathcal{S}$ based on $\mathcal{E}$}{
    let $\mathcal{A}, \mathcal{B}$ be the set of variables in Algorithm \ref{conditionalIndependenceTest} such that $\Sigma_{\mathcal{A}, \mathcal{B}}$ was rank deficient with rank $k$\;
    
    $k_1 \gets \texttt{rank}_{\graphp}(\Sigma_{\mathcal{A}\cup \setL_3, \mathcal{B}})$\;
    
    $k_2 \gets \texttt{rank}_{\graphp}(\Sigma_{\mathcal{A}, \mathcal{B}\cup \setL_3})$\;
    
    \If{$k<\min(k_1, k_2)$}{
        add collider $\setL_1 \rightarrow \setL_3 \leftarrow \setL_2$ to $\mathcal{C}$\;
    }
}
\Return{Collider set $\mathcal{C}$}
\caption{findColliders}
\label{Alg:findColliders}
\end{algorithm}

As mentioned above, we only need to perform cross-cover test and v-structure test locally in the estimated graph $\graphp$. Algorithm \ref{refineEdges} (Phase III: \textit{refineEdges}) combines the search procedure of the two components together with the output $\graph'$ from \textit{refineClusters} as the input. Specifically, for each latent cover $\setL'$ in $\graphp$, we only need to consider testing for edges amongst $\setL'$ and its children $\setC = \childrenp(\setL')$. Thus, we perform a depth-first traversal of the output graph $\graphp$ starting from $Root(\graphp)$ (lines 1-3), and apply the cross-cover test at each $\setL$ to refine the skeleton of $\graphp$ (lines 4-7, 11) and the collider test to identify the v structures (lines 10, 11). After determining v-structures, we can find more directions by applying Meek's rule (line 12), analogous to that in the PC algorithm \citep{spirtes2000causation}.

\begin{algorithm}[!htp]
\SetKwInOut{Input}{Input}
\SetKwInOut{Output}{Output}
\SetKwFunction{refineEdges}{\texttt{refineEdges}}
\Input{Learned graph $\graphp$ from phase II}
\Output{Markov equivalence class $\graphp$}

\SetInd{0.5em}{0.5em}
\Indm\nonl\refineEdges($\graphp$):\\
\Indp
\SetInd{0.5em}{1.0em}
\ForEach{latent atomic cover $\setL'$ in $\graphp$}{
\lIf{$\setL'$ does not have latent children}{
    \Return{$\graphp$}
} 
\lForEach{latent child $\setC_i$ of $\setL'$}{
    $\graphp \gets \texttt{refineEdges}(\graphp, \setC_i)$ 
}
\lIf{$\setC \coloneqq \purechildrenp(L')$ is a single latent cover}{
    $\mathcal{S} \gets \setL' \cup \setC \cup \purechildrenp(\setC)$
}
\lElse{
    $\mathcal{S} \gets \setL' \cup \setC$
}
$\mathcal{G''} \gets \texttt{makeRoot}_{\graphp}(\setL')$ and remove all edges amongst $\mathcal{S}$ in $\mathcal{G''}$ \tcp*{temp graph} 
Edgeset $\mathcal{E} \gets \texttt{crossCoverTest}(\mathcal{S}, \mathcal{G''})$\;

\lIf{no conditional independencies found}{\Return{$\graphp$}}
\Else{
    Collider set $\mathcal{C} \gets \texttt{findColliders}(\mathcal{S}, \mathcal{E}, \mathcal{G''})$\;
    
    in $\graphp$, remove all edges amongst $\mathcal{S}$ and use $\mathcal{E}, \mathcal{C}$ to connect variables in $\mathcal{S}$\;
    
    apply Meek's rule to $\graphp$\;
}
}
   convert $\mathcal{G'}$ to its Markov equivalence class\;
    \Return{$\mathcal{G'}$
}

\caption{Phase III: refineEdges}
\label{refineEdges}
\end{algorithm}

\section{Theoretical Results}
\label{Sec:theory}

In this section, we show the correctness of the algorithms proposed in Section \ref{Sec:Alg}. In particular, by making use of the rank constraints of only the measured variables, the proposed algorithms output the correct Markov equivalence class of the IL$^2$H graph asymptotically, under the \textit{minimal-graph operator} and \textit{skeleton operator}, with their definitions given below. 

\begin{definition}[Markov Equivalence Class of IL$^2$H graphs]
Two IL$^2$H graphs $\graph_1$ and $\graph_2$ are in the same Markov equivalence class, denoted by $\graph_1 \approx \graph_2$, if and only (1) they have the same set of variables (both measured and latent variables), (2) have the same causal skeleton, and (3) have the same V-structures $\setL_i \rightarrow \setL_k \leftarrow \setL_j$, where $\setL_i, \setL_j, \setL_k$ represent latent atomic covers. 
\label{The: equ_class}
\end{definition}

\begin{definition}[Minimal-Graph Operator]
Suppose $\graph$ is an IL$^2$H graph. For every latent atomic cover $\setL$ in $\graph$, merge $\setL$ to its parents $\setP$ if the following conditions hold: (1) $\setL$ is the pure children of $\setP$, (2) $|\setL| = |\setP|$, and (3) the pure children of $\setL$ form one latent atomic cover, or the siblings of $\setL$ form one latent atomic cover. 
We call such operator the minimal-graph operator and denote it by $\mathcal{O}_{\text{min}}(\graph)$.
\end{definition}

\begin{definition}[Skeleton Operator]
  Suppose $\graph$ is an IL$^2$H graph. A skeleton operator of $\graph$, denoted by $\mathcal{O}_s(\graph)$ is defined as follows: for any latent atomic cover $\setL$, draw an edge from $l_j \in \setL$ to $c_k \in PCh_{\graph}(\setL_i)$, if $l_j$ and $c_k$ are not directly connected in $\graph$.
\end{definition}
Note that the minimal-graph operator and the skeleton operator will not change the rank constraints, or in other words, graphs before and after applying the operators are indistinguishable with rank constraints. This result is shown in the following lemma. 

\begin{lemma}
 Suppose $\graph$ is an IL$^2$H graph. The rank constraints are invariant with the \textit{minimal-graph operator} and the \textit{skeleton operator}; that is, $\graph$ and $\mathcal{O}_{skeleton}(\mathcal{O}_{min}(\graph))$ are rank equivalent.
\end{lemma}
These two operators have already been achieved in Algorithm \ref{overallAlg}, so the output of the algorithm is the rank-equivalent graph with the minimal number of latent atomic covers. 

\begin{figure}[htp]
    \centering
    \subfigure[An IL$^2$H graph.]{
	\label{fig:min-a}
	\begin{tikzpicture}[scale=.6, line width=0.5pt, inner sep=0.2mm, shorten >=.1pt, shorten <=.1pt]
		\draw (-0.5, 4) node(L5)  {{\footnotesize\,$L_5$\,}};
		\draw (-2, 2.5) node(L4) {{\footnotesize\,$L_4$\,}};
		\draw (-1, 2.5) node(X8)  {{\footnotesize\,$X_8$\,}};
		\draw (0, 2.5) node(X9)  {{\footnotesize\,$X_9$\,}};
		\draw (1, 2.5) node(X10)  {{\footnotesize\,$X_{10}$\,}};
		
		\draw (-3, 1) node(L1) {{\footnotesize\,$L_1$\,}};
		\draw (-2, 1) node(L2) {{\footnotesize\,$L_2$\,}};
		\draw (-1, 1) node(L3) {{\footnotesize\,$L_3$\,}};
		
		\draw (-4.5, -0.5) node(X1)  {{\footnotesize\,$X_1$\,}};
		\draw (-3.7, -0.5) node(X2)  {{\footnotesize\,$X_2$\,}};
		\draw (-2.9, -0.5) node(X3)  {{\footnotesize\,$X_3$\,}};
		\draw (-2.1, -0.5) node(X4)  {{\footnotesize\,$X_4$\,}};
		\draw (-1.3, -0.5) node(X5)  {{\footnotesize\,$X_5$\,}};
		\draw (-0.5, -0.5) node(X6)  {{\footnotesize\,$X_6$\,}};
		\draw (0.3, -0.5) node(X7)  {{\footnotesize\,$X_7$\,}};

	   \draw[-latex] (L5) -- (L4);
	   \draw[-latex] (L5) -- (X8);
	   \draw[-latex] (L5) -- (X9);
	   \draw[-latex] (L5) -- (X10);
	   \draw[-latex] (L4) -- (L1);
	   \draw[-latex] (L4) -- (L2);
	   \draw[-latex] (L4) -- (L3);
	   
	   \draw[-latex] (L1) -- (X1);
	   \draw[-latex] (L1) -- (X2);
	   \draw[-latex] (L1) -- (X3);
	   \draw[-latex] (L1) -- (X4);
	   \draw[-latex] (L1) -- (X5);
	   \draw[-latex] (L1) -- (X6);
	   
	   \draw[-latex] (L2) -- (X1);
	   \draw[-latex] (L2) -- (X2);
	   \draw[-latex] (L2) -- (X3);
	   \draw[-latex] (L2) -- (X4);
	   \draw[-latex] (L2) -- (X5);
	   \draw[-latex] (L2) -- (X6);
	   \draw[-latex] (L2) -- (X7);
	   
	   \draw[-latex] (L3) -- (X2);
	   \draw[-latex] (L3) -- (X3);
	   \draw[-latex] (L3) -- (X4);
	   \draw[-latex] (L3) -- (X5);
	   \draw[-latex] (L3) -- (X6);
	   \draw[-latex] (L3) -- (X7);

	\end{tikzpicture}
    }
~
    \subfigure[After applying the minimal-graph operator to the graph in (a), $L_4$ is merged to its parent $L_5$, and the rank constraints do not change.]{
	\label{fig:min-b}
	\begin{tikzpicture}[scale=.6, line width=0.5pt, inner sep=0.2mm, shorten >=.1pt, shorten <=.1pt]
		\draw (-0.5, 4) node(L5)  {{\footnotesize\,$L_5$\,}};
		\draw (-1, 2.5) node(X8)  {{\footnotesize\,$X_8$\,}};
		\draw (0, 2.5) node(X9)  {{\footnotesize\,$X_9$\,}};
		\draw (1, 2.5) node(X10)  {{\footnotesize\,$X_{10}$\,}};
		
		\draw (-4, 2.5) node(L1) {{\footnotesize\,$L_1$\,}};
		\draw (-3, 2.5) node(L2) {{\footnotesize\,$L_2$\,}};
		\draw (-2, 2.5) node(L3) {{\footnotesize\,$L_3$\,}};
		
		\draw (-4.5, 1) node(X1)  {{\footnotesize\,$X_1$\,}};
		\draw (-3.7, 1) node(X2)  {{\footnotesize\,$X_2$\,}};
		\draw (-2.9, 1) node(X3)  {{\footnotesize\,$X_3$\,}};
		\draw (-2.1, 1) node(X4)  {{\footnotesize\,$X_4$\,}};
		\draw (-1.3, 1) node(X5)  {{\footnotesize\,$X_5$\,}};
		\draw (-0.5, 1) node(X6)  {{\footnotesize\,$X_6$\,}};
		\draw (0.3, 1) node(X7)  {{\footnotesize\,$X_7$\,}};

	   \draw[-latex] (L5) -- (X8);
	   \draw[-latex] (L5) -- (X9);
	   \draw[-latex] (L5) -- (X10);
	   \draw[-latex] (L5) -- (L1);
	   \draw[-latex] (L5) -- (L2);
	   \draw[-latex] (L5) -- (L3);
	   
	   \draw[-latex] (L1) -- (X1);
	   \draw[-latex] (L1) -- (X2);
	   \draw[-latex] (L1) -- (X3);
	   \draw[-latex] (L1) -- (X4);
	   \draw[-latex] (L1) -- (X5);
	   \draw[-latex] (L1) -- (X6);
	   
	   \draw[-latex] (L2) -- (X1);
	   \draw[-latex] (L2) -- (X2);
	   \draw[-latex] (L2) -- (X3);
	   \draw[-latex] (L2) -- (X4);
	   \draw[-latex] (L2) -- (X5);
	   \draw[-latex] (L2) -- (X6);
	   \draw[-latex] (L2) -- (X7);
	   
	   \draw[-latex] (L3) -- (X2);
	   \draw[-latex] (L3) -- (X3);
	   \draw[-latex] (L3) -- (X4);
	   \draw[-latex] (L3) -- (X5);
	   \draw[-latex] (L3) -- (X6);
	   \draw[-latex] (L3) -- (X7);

	\end{tikzpicture}
    }
 ~
    \subfigure[After applying the skeleton operator to the graph in (b), $L_1$ has an edge to $X_7$ and $L_3$ has an edge to $X_1$, and the rank constraints do not change.]{
	\label{fig:min-c}
	\begin{tikzpicture}[scale=.6, line width=0.5pt, inner sep=0.2mm, shorten >=.1pt, shorten <=.1pt]
		\draw (-0.5, 4) node(L5)  {{\footnotesize\,$L_5$\,}};
		\draw (-1, 2.5) node(X8)  {{\footnotesize\,$X_8$\,}};
		\draw (0, 2.5) node(X9)  {{\footnotesize\,$X_9$\,}};
		\draw (1, 2.5) node(X10)  {{\footnotesize\,$X_{10}$\,}};
		
		\draw (-4, 2.5) node(L1) {{\footnotesize\,$L_1$\,}};
		\draw (-3, 2.5) node(L2) {{\footnotesize\,$L_2$\,}};
		\draw (-2, 2.5) node(L3) {{\footnotesize\,$L_3$\,}};
		
		\draw (-4.5, 1) node(X1)  {{\footnotesize\,$X_1$\,}};
		\draw (-3.7, 1) node(X2)  {{\footnotesize\,$X_2$\,}};
		\draw (-2.9, 1) node(X3)  {{\footnotesize\,$X_3$\,}};
		\draw (-2.1, 1) node(X4)  {{\footnotesize\,$X_4$\,}};
		\draw (-1.3, 1) node(X5)  {{\footnotesize\,$X_5$\,}};
		\draw (-0.5, 1) node(X6)  {{\footnotesize\,$X_6$\,}};
		\draw (0.3, 1) node(X7)  {{\footnotesize\,$X_7$\,}};

	   \draw[-latex] (L5) -- (X8);
	   \draw[-latex] (L5) -- (X9);
	   \draw[-latex] (L5) -- (X10);
	   \draw[-latex] (L5) -- (L1);
	   \draw[-latex] (L5) -- (L2);
	   \draw[-latex] (L5) -- (L3);
	   
	   \draw[-latex] (L1) -- (X1);
	   \draw[-latex] (L1) -- (X2);
	   \draw[-latex] (L1) -- (X3);
	   \draw[-latex] (L1) -- (X4);
	   \draw[-latex] (L1) -- (X5);
	   \draw[-latex] (L1) -- (X6);
	   \draw[-latex] (L1) -- (X7);
	   
	   \draw[-latex] (L2) -- (X1);
	   \draw[-latex] (L2) -- (X2);
	   \draw[-latex] (L2) -- (X3);
	   \draw[-latex] (L2) -- (X4);
	   \draw[-latex] (L2) -- (X5);
	   \draw[-latex] (L2) -- (X6);
	   \draw[-latex] (L2) -- (X7);
	   
	   \draw[-latex] (L3) -- (X1);
	   \draw[-latex] (L3) -- (X2);
	   \draw[-latex] (L3) -- (X3);
	   \draw[-latex] (L3) -- (X4);
	   \draw[-latex] (L3) -- (X5);
	   \draw[-latex] (L3) -- (X6);
	   \draw[-latex] (L3) -- (X7);

	\end{tikzpicture}
    }
	\caption{Examples of applying the minimal-graph operator and the skeleton operator to an IL$^2$H graph.}
	\label{fig:minial graph}
\end{figure}
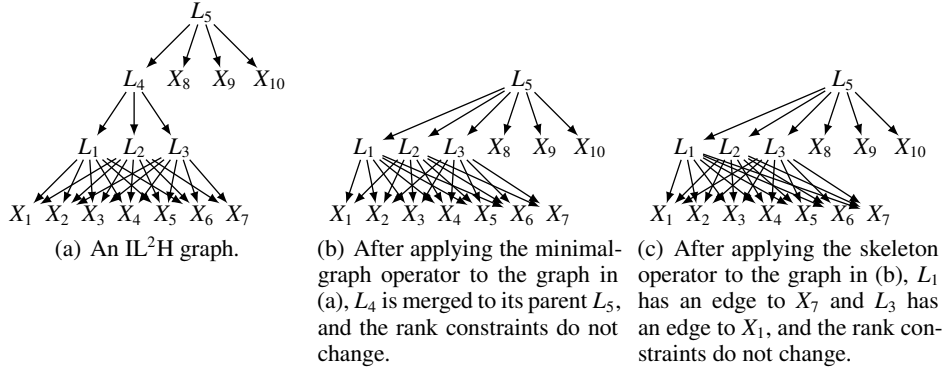

\begin{example}
 Give an IL$^2$H graph in Figure \ref{fig:min-a}, after applying the minimal-graph operator, the latent atomic cover $L_4$ will be merged to its parent $L_5$, resulting in the graph in Figure \ref{fig:min-b}, while the rank constraints will not change. Furthermore, after applying the skeleton operator to the graph in Figure \ref{fig:min-b}, $L_1$ has an edge to $X_7$ and $L_3$ has an edge to $X_1$, resulting in the graph in Figure \ref{fig:min-c}, which also does not change the rank constaints.
\end{example}

We next proceed to show that phases I-III will output a graph $\graphp$ such that $\graphp$ will be in the same Markov equivalence class as $\mathcal{O}_{min}(\mathcal{O}_s(\graph))$, denoted by $\graphp \approx \graph$. We have already shown earlier in Theorem \ref{The:correct_cluster} that \textit{findCausalClusters} gives correct latent covers when there is no bond set. Furthermore, Theorem \ref{The:remove_bond} shows that even in the presence of bond sets, Phase II \textit{refineClusters} is able to refine clusters into those without bond sets. Therefore, Phase I-II can correctly identify the latent atomic covers of $\mathcal{O}_{min}(\graph)$, which is given in the following theorem. 
\begin{theorem}[Identifiability of Latent Variables]
Suppose $\graph$ is an IL$^2$H graph with measured variables $\measures$. Phases I-II in Algorithm \ref{overallAlg} over $\measures$ can asymptotically identify the latent atomic covers of $\mathcal{O}_{min}(\graph)$, with only the first two conditions in Condition 1.
\label{The: identifiability1}
\end{theorem}

Moreover, Phase III \textit{refineEdges} further guarantees correct skeletons and v structures. Therefore, the following theorem shows that Algorithm \ref{overallAlg}, which includes Phases I-III, can asymptotically identify the Markov equivalence class, up to the skeleton operator and the minimal-graph operator.

\begin{theorem}[Identifiability of Causal Graph]
Suppose $\graph$ is an IL$^2$H graph with measured variables $\measures$. Algorithm \ref{overallAlg}, including Phases I-III, over $\measures$ can asymptotically identify the Markov equivalence class of $\mathcal{O}_{min}(\mathcal{O}_{s}(\graph))$.
\label{The: identifiability2}
\end{theorem}

\section{Experimental Results} \label{Sec:exp}
We applied the proposed algorithm to synthetic data to learn the latent hierarchical causal graph. Specifically, we considered different types of latent graphs and different sample sizes (with $N\!=\!2k, 5k, 10k$). The causal strength was generated uniformly from $[-5, -0.5] \cup [0.5, 5]$, and the noise term either follows a Gaussian distribution (with noise variance uniformly sampled from $[1, 5]$) or a uniform distribution $\mathcal{U}(-2, 2)$.

To the best of our knowledge, this is the first algorithm that can identify such general latent hierarchical structures, so to fairly compare with other methods, besides general IL$^2$H graphs (see Figure \ref{fig:IL2H}), we also considered tree structures (see Figure \ref{fig:tree}) and measurement models (see Figure \ref{fig: measurement}). We compared the proposed method with the tree-based method--Chow-Liu Recursive Grouping (CLRG) \citep{choi2011latenttree}, as well as measurement-model-based methods, including FOFC \citep{Kummerfeld2016} and GIN \cite{xie2020generalized}.

We used the following metrics to evaluate the performance:
\begin{itemize}[leftmargin=*,align=left,itemsep=0pt,topsep=0pt]
    \item \textit{Causal cluster recovery rate over measured variables (metric 1)}: measured by the percentage of correctly identified causal clusters over measured variables, with $$\textit{m}_1 = \frac{\text{correctly found \# clusters over measured variables}}{\text{total \# clusters over measured variables}}.$$
    \item \textit{Causal cluster recovery rate over all variables (metric 2)}, measured by the percentage of correctly identified causal clusters over all variables, with $$\textit{m}_2 = \frac{\text{correctly found \# clusters over all variables}}{\text{total \# clusters over all variables}}.$$
    \item  \textit{Percentage differences between estimated and true adjacency matrices (metric 3)}, with $$\textit{m}_3 = \sum_{i,j} \big(\textit{Adj}_{\graph}(i,j) \sim= \textit{Adj}_{\graphp}(i,j)\big)/\big((n_\setX + n_\setL)^2-n_\setX^2\big),$$ where $\textit{Adj}$ denotes the adjacency matrix, $i$ and $j$ denote the $i$-th and $j$-th entry, respectively, and $n_\setX$ and $n_\setL$ are the number of measured variables and latent variables, respectively. 
    Note that the indices of the latent variables in the estimated graph may not be aligned to those in the true graph. To remove this ambiguity, we tried all permutations of the latent indices in the estimated graph and used the one which has the smallest difference from the true graph. Moreover, if the estimated number of latent variables is smaller than the true number of latent variables, add extra latent variables to $\graphp$ that do not have edges with others. If the estimated number of latent variables is larger than the true number of latent variables, then find a subset of the latent variables in $\graphp$ that best aligns the true ones. 
\end{itemize}
It is worth mentioning that how to measure the performance of the estimated latent hierarchical graph is a nontrivial problem and will be further investigated. 

The experimental results were reported in Tables \ref{tab:experimental_compare_gaussian} and \ref{tab:experimental_compare_uniform}, where the noise terms are Gaussian distributed and uniformly distributed, respectively. Our method gives the best results on all types of graphs, indicating that it can handle not only the tree-based and measurement-based structures, but also the latent hierarchical structure. The CLRG algorithm does not perform well on tree-based structure because the first two metrics are rather strict--even a single mis-clustered variable outputs an error.

\begin{center}
 \begin{table*}[hpt]
	\small
	\center \caption{Performance (mean (standard deviation)) on learning different types of latent graphs, where noise terms were generated from Gaussian distributions.}
	\label{tab:experimental_compare_gaussian}
	\begin{center}
\begin{tabular}{|c|c|c|c|c|c|}
		\hline  \multicolumn{2}{|c|}{} &\multicolumn{4}{|c|}{\textbf{metric 1 $\uparrow$}}\\
		\hline 
		\multicolumn{2}{|c|}{Algorithm} & Ours & CLRG  & FOFC & GIN \\
		\hline 
		 & 2k 
		 & 0.70 (0.22) & 0.00 (0.00) & 0.12 (0.09) & 0.35 (0.23) \\
		\cline{2-6}
		{\emph{IL$^2$H}}
		 &5k 
		 & 0.83 (0.15) & 0.00 (0.00) & 0.17 (0.10) & 0.40 (0.24)  \\
		\cline{2-6}
		 &10k
		 & 0.86 (0.13) & 0.00 (0.00) & 0.29 (0.10) & 0.44 (0.21)  \\
		\hline 
		 & 2k 
		 & 0.89 (0.12) & 0.00 (0.00) & 0.38 (0.25) & 0.13 (0.30) \\
		 \cline{2-6}
		 {\emph{Tree}}
		 &5k 
		 & 1.0 (0.00) & 0.00 (0.00) & 0.75 (0.23) & 0.23 (0.30)  \\
		\cline{2-6}
		 &10k
		 & 1.0 (0.00) & 0.13 (0.04) & 0.87 (0.20) & 0.50 (0.00) \\
		\hline 
		 & 2k 
		 & 0.92 (0.08) & 0.00 (0.00) & 0.38 (0.22) & 0.30 (0.16) \\
		\cline{2-6}
		{\emph{Measurement  Model}} &5k 
		& 1.0 (0.00) & 0.00 (0.00) & 0.65 (0.32) & 0.70 (0.43) \\
		\cline{2-6}&10k 
		 & 1.0 (0.00) & 0.00 (0.00) & 1.0 (0.00) & 0.70 (0.43) \\
		\hline 
	\end{tabular}
	\end{center}
	\hfill
	\vspace{.5cm}
	\begin{center}
	\begin{tabular}{|c|c|c|c|c|c|}
		\hline  \multicolumn{2}{|c|}{} &\multicolumn{4}{|c|}{\textbf{metric 2 $\uparrow$}}\\
		\hline 
		\multicolumn{2}{|c|}{Algorithm} & Ours & CLRG  & FOFC & GIN \\
		\hline 
		 & 2k 
		 & 0.60 (0.16) & 0.00 (0.00) & 0.09 (0.07) & 0.26 (0.19) \\
		\cline{2-6}
		{\emph{IL$^2$H}}
		 &5k 
		 & 0.69 (0.22) & 0.00 (0.00) & 0.09 (0.07) & 0.28 (0.19)  \\
		\cline{2-6}
		 &10k
		 & 0.73 (0.17) & 0.00 (0.00) & 0.12 (0.07) & 0.35 (0.17)  \\
		\hline 
		 & 2k 
		 & 0.79 (0.19) & 0.00 (0.00) & 0.28 (0.19) & 0.09 (0.11) \\
		 \cline{2-6}
		 {\emph{Tree}}
		 &5k 
		 & 0.83 (0.16) & 0.00 (0.00) & 0.55 (0.17) & 0.17 (0.22) \\
		\cline{2-6}
		 &10k
		 & 0.89 (0.09) & 0.09 (0.03) & 0.63 (0.18) & 0.36 (0.20) \\
		\hline 
		 & 2k 
		 & 0.92 (0.08) & 0.00 (0.00) & 0.38 (0.22) & 0.30 (0.16) \\
		\cline{2-6}
		{\emph{Measurement  Model}} &5k 
		& 1.0 (0.00) & 0.00 (0.00) & 0.65 (0.32) & 0.70 (0.43) \\
		\cline{2-6}&10k 
		 & 1.0 (0.00) & 0.00 (0.00) & 1.0 (0.00) & 0.70 (0.43) \\
		\hline 
	\end{tabular}
	\end{center}
	\hfill
		\vspace{.5cm}
	\begin{center}	
	\begin{tabular}{|c|c|c|c|c|c|}
		\hline  \multicolumn{2}{|c|}{} &\multicolumn{4}{|c|}{\textbf{metric 3 $\downarrow$}} \\
		\hline 
		\multicolumn{2}{|c|}{Algorithm} & Ours & CLRG  & FOFC & GIN  \\
		\hline 
		 & 2k 
		 & 0.11 (0.02) & 0.18 (0.04) & 0.26 (0.21) & 0.16 (0.03)  \\
		\cline{2-6}
		{\emph{IL$^2$H}}
		 &5k 
		 & 0.10 (0.02) & 0.18 (0.04) & 0.26 (0.21) & 0.15 (0.03)  \\
		\cline{2-6}
		 &10k
		 & 0.10 (0.02) & 0.18 (0.04) & 0.20 (0.27) & 0.15 (0.03)  \\
		\hline 
		 & 2k 
		 & 0.02 (0.00) & 0.10 (0.00) & 0.09 (0.00) & 0.15 (0.02) \\
		 \cline{2-6}
		 {\emph{Tree}}
		 &5k 
		 & 0.02 (0.00) & 0.10 (0.00) & 0.09 (0.00) & 0.10 (0.02) \\
		\cline{2-6}
		 &10k
		 & 0.01 (0.00) & 0.09 (0.00) & 0.08 (0.00) & 0.09 (0.00) \\
		\hline 
		 & 2k 
		 & 0.02 (0.00) & 0.28 (0.00) & 0.18 (0.04) & 0.30 (0.08) \\
		\cline{2-6}
		{\emph{Measurement  Model}} &5k 
		& 0.00 (0.00) & 0.28 (0.00) & 0.13 (0.05) & 0.17 (0.15) \\
		\cline{2-6}&10k 
		 & 0.00 (0.00) & 0.28 (0.00) & 0.00 (0.00) & 0.17 (0.15) \\
		\hline 
	\end{tabular}
	\end{center}
	
	\begin{tablenotes}
		\item Note: $\uparrow$ means a higher value is better, and vice versa.
	\end{tablenotes}
 \end{table*}
\end{center}

\begin{center}
\begin{table*}[hpt]
	\small
	\center \caption{Performance (mean (standard deviation)) on learning different types of latent graphs, where noise terms were generated from uniform distributions.}
	\label{tab:experimental_compare_uniform}
	\begin{center}
	\begin{tabular}{|c|c|c|c|c|c|}
		\hline  \multicolumn{2}{|c|}{} &\multicolumn{4}{|c|}{\textbf{metric 1 $\uparrow$}}  \\
		\hline 
		\multicolumn{2}{|c|}{Algorithm} & Ours & CLRG  & FOFC & GIN  \\
		\hline 
		 & 2k 
		 & 0.72 (0.19) & 0.00 (0.00) & 0.10 (0.10) & 0.45 (0.18)   \\
		\cline{2-6}
		{\emph{IL$^2$H}}
		 &5k 
		 & 0.87 (0.13) & 0.00 (0.00) & 0.17 (0.10) & 0.46 (0.18)  \\
		\cline{2-6}
		 &10k
		 & 0.88 (0.10) & 0.00 (0.00) & 0.29 (0.10) & 0.52 (0.16)  \\
		\hline 
		 & 2k 
		 & 0.93 (0.07) & 0.00 (0.00) & 0.40 (0.27) & 0.79 (0.22) \\
		 \cline{2-6}
		 {\emph{Tree}}
		 &5k 
		 & 1.0 (0.00) & 0.00 (0.00) & 0.75 (0.23) & 0.99 (0.04)  \\
		\cline{2-6}
		 &10k
		 & 1.0 (0.00) & 0.00 (0.00) & 0.87 (0.20) & 0.95 (0.08) \\
		\hline 
		 & 2k 
		 & 0.95 (0.06) & 0.00 (0.00) & 0.25 (0.21) & 0.87 (0.20) \\
		\cline{2-6}
		{\emph{Measurement Model}} &5k 
		& 1.0 (0.00) & 0.00 (0.00) & 0.70 (0.27) & 1.00 (0.00) \\
		\cline{2-6}&10k 
		 & 1.0 (0.00) & 0.00 (0.00) & 1.0 (0.00) & 1.00 (0.00) \\
		\hline 
	\end{tabular}
	\end{center}
	\hfill
\vspace{.5cm}
   \begin{center}
	\begin{tabular}{|c|c|c|c|c|c}
		\hline  \multicolumn{2}{|c|}{} &\multicolumn{4}{|c|}{\textbf{metric 2 $\uparrow$}}\\
		\hline 
		\multicolumn{2}{|c|}{Algorithm} & Ours & CLRG  & FOFC & GIN  \\
		\hline 
		 & 2k 
		 & 0.61 (0.20) & 0.00 (0.00) & 0.07 (0.07) & 0.34 (0.15)  \\
		\cline{2-6}
		{\emph{IL$^2$H}}
		 &5k 
		 & 0.71 (0.19) & 0.00 (0.00) & 0.09 (0.07) & 0.35 (0.15)  \\
		\cline{2-6}
		 &10k
		 & 0.76 (0.20) & 0.00 (0.00) & 0.12 (0.07) & 0.39 (0.14)  \\
		\hline 
		 & 2k 
		 & 0.83 (0.12) & 0.00 (0.00) & 0.29 (0.20) & 0.57 (0.16) \\
		 \cline{2-6}
		 {\emph{Tree}}
		 &5k 
		 & 0.85 (0.11) & 0.00 (0.00) & 0.55 (0.17) & 0.72 (0.03) \\
		\cline{2-6}
		 &10k
		 & 0.89 (0.15) & 0.00 (0.00) & 0.63 (0.18) & 0.69 (0.06) \\
		\hline 
		 & 2k 
		 & 0.95 (0.06) & 0.00 (0.00) & 0.25 (0.21) & 0.87 (0.20) \\
		\cline{2-6}
		{\emph{Measurement Model}} &5k 
		& 1.0 (0.00) & 0.00 (0.00) & 0.70 (0.27) & 1.00 (0.00) \\
		\cline{2-6}&10k 
		 & 1.0 (0.00) & 0.00 (0.00) & 1.0 (0.00) & 1.00 (0.00) \\
		\hline 
	\end{tabular}
	\end{center}
	
	\hfill
	\vspace{0.5cm}
	\begin{center}
	\begin{tabular}{|c|c|c|c|c|c|}
		\hline  \multicolumn{2}{|c|}{} &\multicolumn{4}{|c|}{\textbf{metric 3 $\downarrow$}} \\
		\hline 
		\multicolumn{2}{|c|}{Algorithm} & Ours & CLRG  & FOFC & GIN  \\
		\hline 
		 & 2k 
		 & 0.10 (0.02) & 0.18 (0.04) & 0.23 (0.21) & 0.15 (0.03)  \\
		\cline{2-6}
		{\emph{IL$^2$H}}
		 &5k 
         & 0.10 (0.02) & 0.18 (0.04) & 0.26 (0.21) & 0.15 (0.03)  \\
		\cline{2-6}
		 &10k
		 & 0.10 (0.02) & 0.18 (0.04) & 0.20 (0.27) & 0.14 (0.03)  \\
		\hline 
		 & 2k 
		 & 0.02 (0.00) & 0.11 (0.00) & 0.10 (0.00) & 0.11 (0.00) \\
		 \cline{2-6}
		 {\emph{Tree}}
		 &5k 
		 & 0.01 (0.00) & 0.10 (0.00) & 0.09 (0.00) & 0.11 (0.00) \\
		\cline{2-6}
		 &10k
		 & 0.01 (0.00) & 0.10 (0.00) & 0.08 (0.00) & 0.11 (0.00) \\
		\hline 
		 & 2k 
		 & 0.01 (0.00) & 0.28 (0.00) & 0.25 (0.04) & 0.07 (0.05) \\
		\cline{2-6}
		{\emph{Measurement Model}} &5k 
		& 0.00 (0.00) & 0.28 (0.00) & 0.25 (0.04) & 0.07 (0.05) \\
		\cline{2-6}&10k 
		 & 0.00 (0.00) & 0.28 (0.00) & 0.00 (0.00) & 0.00 (0.00) \\
		\hline 
	\end{tabular}
	\end{center}
	\begin{tablenotes}
		\item Note: $\uparrow$ means a higher value is better, and vice versa.
	\end{tablenotes}
\end{table*}
\end{center}

\begin{figure}[htp]
    \begin{minipage}{0.45\textwidth}
     \centering
	%
	\caption{Tree. Note that $X_{i,j,k}$ means $X_i,X_j,X_k$.}
	\label{fig:tree}
	\end{minipage}\hfill
   \begin{minipage}{0.48\textwidth}
\centering
    %
    \caption{Measurement model.}
    \label{fig: measurement}
    \end{minipage}
\end{figure}

\begin{figure}[htp!]
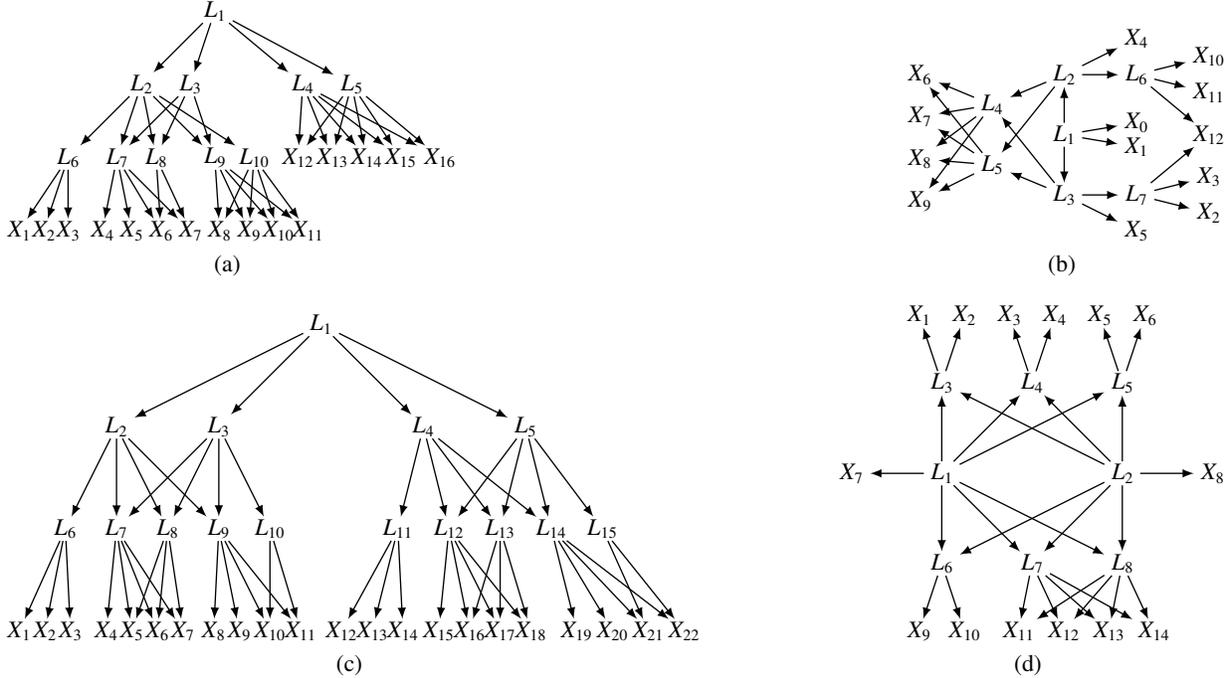

    \centering
    \subfigure[]{
	%
    }
	\caption{Example IL$^2$H graphs.}
	\label{fig:IL2H}
\end{figure}

\section{Conclusions and Future Work}
In this paper, we formulated a specific type of latent hierarchical causal model and proposed a method to identify its graph by making use of rank deficiency constraints. Theoretically, we show that the proposed algorithm can find the correct Markov equivalence class of the whole graph asymptotically under mild restrictions of the graph structure. For more general graphs, only using the second-order statistics may result in a rank equivalence class that contains multiple DAGs, so how to further leverage high-order statistics to distinguish between causal graphs within the equivalence class will be our future work. Other future research directions include allowing nonlinear causal relationships and allowing measured variables to cause latent variables (existing techniques, e.g., \citet{adams2021identification, Chandler22}, may help to mitigate this issue).

\section{Acknowledgement}
BH would like to acknowledge the support of Apple Scholarship. KZ would like to acknowledge the support by the National Institutes of Health (NIH) under Contract R01HL159805, by the NSF-Convergence Accelerator Track-D award \#2134901, and by a grant from Apple and KDDI.

\appendix

\section{Appendix}

\subsection{Related Work} 
Identification of causal relationships from observational data, known as \textit{causal discovery}, is attractive for the reason that traditional randomized control trials may be hard or even impossible to do. Most state-of-the-art approaches in causal discovery assume that the measured variables are the underlying causal variables and that no latent confounders influence the measured variables \citep{spirtes2000causation,chickering2002optimal,lingam,hoyer2009ANM,zhang2009PNL}. However, in many real-world problems, this assumption may not hold. 

For example, in complex systems, it is usually hard to enumerate and measure all task-related variables, so there may exist latent variables that influence multiple observed variables, the ignorance of which may introduce spurious correlations between measured variables. A more complex scenario is that the variables form a hierarchical structure, where the latent variables may generate latent variables in a hierarchical way, while only the leaf nodes are measured, which is common in real-world scenarios. For instance, in fMRI data analysis, hundreds of thousands of voxels are recorded, where these micro-variables may not be necessary to have clear semantic meaning. Therefore, from the measured voxels, we aim to automatically identify conceptually meaningful functional brain regions of different levels, where the lower level represents simpler functional regions and the higher level represents more abstract and complex functional regions, which thus form a hierarchical structure. We may also see similar structures in image representation learning--image pixels are dependent, and it seems sensible to consider them as observations generated by multiple-layer hidden concepts.

Previous causal discovery approaches that can handle latent confounders are mainly based on the following criteria.
\begin{itemize}[leftmargin=*,align=left,itemsep=0pt,topsep=-1pt]
    \item Conditional independence constraints. The FCI algorithm \citep{spirtes2000causation}, as well as its variants \citep{colombo2012learning,Pearl:2000:CMR:331969,Akbari2021}, makes use of conditional independence tests over observed variables to identify the causal structure over observed variables up to a maximal ancestral graph. This type of methods can handle both linear and nonlinear causal relationships, but the limitation is that there are large indeterminacies in the resulting graph about the existence of an edge, as well as the existence of confounders. In practice, it is often the case that the resulting graph contains many undetermined edges, denoted by $\multimapboth$, where the circle can be either tail or arrow. Moreover, they do not consider the causal relationships among latent variables.
    \item Tetrad condition. With the Tetrad condition, i.e., the rank constraints of every $2 \times 2$ off-diagonal sub-covariance matrix, one is able to locate latent variables and identify the causal skeleton among them in linear-Gaussian models \citep{Silva-linearlvModel, Kummerfeld2016, Shuyan20}. These methods assume that each observed variable is influenced by only one latent parent, and each latent variable has at least three pure measured children. Moreover, the Tetrad condition can also be used to identify a latent tree structure \citep{Pearl1988}. 
    \item Matrix decomposition. It has been shown that, under certain conditions, the precision matrix can be decomposed into a low-rank matrix and a sparse matrix, where the low-rank matrix represents the causal structure from latent variables to observed variables and the sparse matrix gives the structural relationships over observed variable. To achieve such decomposition, however, certain assumptions are imposed on the structure \citep{RankSparsity_11, RankSparsity_12}. A related work \citep{anandkumar2013learning} decomposed the covariance matrix into a low-rank matrix and a diagonal matrix, which requires three times more measured variables than latent variables. 
    \citep{anandkumar2013learning} can also handle multi-level DAGs that some latent variables do not have measured variables as children, but it requires that the underlying graph can be partitioned into multiple levels such that all the edges are between nodes in adjacent layers; the graphs in Figure 1(b) in the main text, and Figures 5-6 and Figures 7(a, b, d) in Appendix are not satisfied.
    \item Over-complete independent component analysis (ICA)-based methods. Several methods \citep{shimizu2009estimation} make use of over-complete ICA to learn the causal structure with latent variables, since it allows more source signals than observed variables. These methods do not consider the causal structure among latent variables and the size of the equivalence class of the identified structure could be  large. In addition, in practice, the estimation of over-complete ICA models is easy to get stuck in local optima, unless the underlying sources are very sparse \citep{entner2010discovering, tashiro2014parcelingam}.
    \item Generalized independent noise (GIN) condition. The GIN condition is an extension of the independent noise condition in the existence of latent confounders. It assumes the noise terms are non-Gaussian and leverages higher-order statistics to identify latent structures. In particular, \citet{xie2020generalized} proposes a GIN-based approach that allows multiple latent parents behind every pair of observed variables and can identify causal directions among latent variables, but it still requires that each latent variable set should have at least twice more measured variables as children.
    \item Moreover, \citet{CDNOD_jmlr} considered a special type of confounders in heterogeneous data, where the confounder can be represented as a function of domain index or a smooth function of time, so one may use the known domain index or time index as a surrogate variable to remove the influence from those confounders and thus identify causal structure over observed variables.
    \item Mixture oracles-based method. Recently, \citet{kivva2021learning} proposed a mixture oracles-based method to identify the latent variable graph that allows nonlinear causal relationships. It is based on assumptions that the latent variables are discrete and each latent variable has measured variables as children. Thanks to the discreteness assumption, it can handle more general DAGs over latent variables.
\end{itemize}

\newtheorem{innercustomthm}{Theorem}
\newenvironment{customthm}[1]
  {\renewcommand\theinnercustomthm{#1}\innercustomthm}
  {\endinnercustomthm}
  
\newtheorem{innercustomlem}{Lemma}
\newenvironment{customlem}[1]
  {\renewcommand\theinnercustomlem{#1}\innercustomlem}
  {\endinnercustomlem}
  
\newtheorem{innercustomprop}{Proposition}
\newenvironment{customprop}[1]
  {\renewcommand\theinnercustomprop{#1}\innercustomprop}
  {\endinnercustomprop}
  
\newtheorem{innercustomcor}{Corollary}
\newenvironment{customcor}[1]
  {\renewcommand\theinnercustomcor{#1}\innercustomcor}
  {\endinnercustomcor}

\subsection{Proof of Theorem 1}
\begin{customthm}{1}[Graphical Implication of Rank Constraints in IL$^2$H Graphs]
Suppose $\mathcal{G}$ satisfies an IL$^2$H graph. Under the rank faithfulness assumption, the cross-covariance matrix $\Sigma_{\setX_A,\setX_B}$ over measured variables $\setX_A$ and $\setX_B$ in $\mathcal{G}$ (with $|\setX_A|, |\setX_B|>r$) has rank $r$, if and only if there exists a subset of latent variables $\setL$ with $|\setL| = r$ such that $\setL$ d-separates $\setX_A$ from $\setX_B$, and there is no $\mathbf{L}'$ with $|\mathbf{L}'| < |\setL|$ that d-separates $\setX_A$ from $\setX_B$. That is,
$$rank(\Sigma_{\setX_A,\setX_B}) = min \{ |\setL|: \setL \text{ d-separates } \setX_A \text{ from } \setX_B \}.$$
\end{customthm}

\begin{proof}
Theorem 1 is a special case of Theorem 2.8 in \citet{tseparation} when applied to IL$^2$H graphs. Different from the setting in \citet{tseparation} which access to the full covariance matrix $\Sigma_{\variables,\variables}$ is assumed, we only have access to the covariance matrix $\Sigma_{\measures,\measures}$ over the measured variables $\measures$. 

It is enough to show that for IL$^2$H graphs, $(C_A, C_B)$ t-separating $\setX_A$ from $\setX_B$ is equivalent to $\setL$ d-separating $\setX_A$ from $\setX_B$, where $C_A, C_B \subset \setL_{\graph}$. 

Since $\setL$ d-separates $\setX_A$ from $\setX_B$ and since any $X \in \setX_{\graph}$ cannot be the choke point, we can choose $C_A=\setL$ and $C_B = \emptyset$, so that $(C_A, C_B)$ t-separates $\setX_A$ from $\setX_B$ in IL$^2$H graphs. 

Therefore, combining Theorem 2.8 in \citet{tseparation} and the above equivalence, the theorem is proved. 
\end{proof}

\subsection{Proof of Theorem 2}
\begin{customthm}{2}[Measurement as a surrogate]
 Suppose $\graph$ is an IL$^2$H graph. Denote by $\setA, \setB \subseteq \setV_{\graph}$ two subsets of variables in $\graph$, with $\setA \cap \setB = \emptyset$. Furthermore, denote by $\setX_A$ the set of measured variables that are d-separated by $\setA$ from all other measures, and by $\setX_B$ the set of measured variables that are d-separated by $\setB$ from all other measures. Then $\texttt{rank}(\Sigma_{\setA,\setB}) = \texttt{rank}(\Sigma_{\setX_A,\setX_B}).$
\end{customthm}

\begin{proof}
According to Theorem 1, the rank of $\Sigma_{X_\setA,X_\setB}$ is the minimal number of latent variables $\setL$ that d-separate $X_\setA$ and $X_\setB$; that is, $\setL$ block all paths between $X_\setA$ and $X_\setB$ with the smallest cardinality. 
Furthermore, since $\setA$ d-separates $\setX_A$ from all other measures, $\setX_A$ are the measured pure descendants of $\setA$. Similarly, $\setX_B$ are the measured pure descendants of $\setB$. Moreover, given the structure of IL$^2$H graphs, $\setL$ also block all paths between $\setA$ and $\setB$ with the smallest cardinality. Therefore, the rank of $\Sigma_{\setA,\setB}$ is also $|\setL|$, equivalent to that of $\Sigma_{\setX_A,\setX_B}$.
\end{proof}

\subsection{Proof of Theorem 3}
\begin{customthm}{3}[Correct Cluster Condition]
Suppose $\graph$ is an IL$^2$H graph with measured variables $\measures$. Consider the output $\graphp$ from applying \textit{findCausalClusters} over $\measures$. If none of the clusters in $\graphp$ is the bond set in $\graph$, then all latent atomic covers have been correctly identified. 
\end{customthm}

Before the proof of Theorem 3, we first give the following lemma which shows that if $\setX \subseteq \measures$ is not a correct cluster in the true graph $\graph$, then $\setX$ forms a bond set. In other words, If a set of measured variables does not form a bond set of $\graph$, then it must form a correct cluster.
\begin{lemma}[Fake-Cluster $\Rightarrow$ Bond Set]
A set of measured variables $\setX \subseteq \measures$ that is mistakely tested as a rank-deficient set by \textit{findCausalClusters} but does not form a cluster in the true graph $\graph$ (that is, a fake cluster), only if $\setX$ forms a bond set of $\graph$.
\label{The:fake-cluster}
\end{lemma}
Below, we first give the proof of Lemma \ref{The:fake-cluster}.
\begin{proof}
Suppose we found a rank deficient set of variables $\setX = \{X_1, X_2, \cdots\}$ which is not a cluster in $\graph$. This implies that (i) there exists at least two disjoint latent atomic covers, $\setL_1, \setL_2$, which d-separate variables in $\setX$ from all other measures, and (ii) $\setX$ does not contain all the pure children of either $\setL_1$ nor $\setL_2$, because otherwise, due to the IL$^2$H requirement that each $\setL_i$ has $> |\setL_i|$ pure children, we would have discovered that a subset of $\setX$ was rank deficient and clustered together earlier by \textit{findCausalClusters}. This implies that $\setL_1, \setL_2$ will d-separate the remaining pure child of $\setL_1$ from that of $\setL_2$, implying that $\setX$ is a bond set.
\end{proof}

Now we are ready to prove Theorem 3.
\begin{proof}
We will prove that all latent atomic covers can be correctly identified \textit{from bottom to top}, if there is no bond set. 

First, from Lemma \ref{The:fake-cluster} we know that if a set of measured variables is not a bond set, then they form a correct causal cluster, and thus the corresponding identified latent cover is correct. Denote the latent atomic covers identified at this step by $\setL_1'$, and denote by $\graph'_1$ the currently estimated graph. 

After identifying the latent atomic covers at the downmost level, we next continue to form causal clusters from root variables in $\graph'_1$, including $\setL_1'$ and the remaining measured variables that did not form clusters in the previous step. If any of the latent covers in $\setL_1'$ have latent children $\setL_i$ in the true graph $\graph$ that have not been identified in the current step, then reverse these edges such that the children become parents. Such an operation will not affect the discovery of latent covers, and $\setL_i$ will be found in later steps. So by further leveraging Lemma \ref{The:fake-cluster} and by treating $\setL_1'$ and the remaining measured variables that did not form clusters in the previous step as ``measured variables", this step results in correctly identified latent covers $\setL_2'$ and estimated graph $\graph_2'$.

We can now iteratively repeat the previous step to discover new latent atomic covers from the root variables in the estimated graph in the previous step, until no more rank deficient sets can be found.
\end{proof}

\subsection{Proof of Theorem 4}
\begin{customthm}{4}[Correcting Clusters]
Denote by $\graphp$ the output from \textit{findCausalClusters} and by $\graph$ the true graph. For a latent atomic cover $\setL'$ in $\graphp$, if the measured pure descendants of $\setL'$ is a bond set in the true graph $\graph$, then there exist a set of siblings $\setS$ of $\setL'$ in $\graphp$, a set of children $\setC$ of $\setL'$, and a set of grandparents $\setP$ of $\setL'$, such that $\measuredp(\setS \cup \setC \cup \setP)$ forms a cluster that is not a bond set in $\graph$.
\end{customthm}

\begin{proof}
Suppose the measured pure descendants of $\setL'$, $\setX \coloneqq \mathcal{M}_{\graphp} (\setL')$, is a bond set in the true graph $\graph$. Denote by $\setL_S \subseteq \latents$ the minimal set of latent variables in $\graph$ that d-separates $\setX$ from all other measures $\setX' \coloneqq \measures \backslash \setX$, and since $\setX$ is a bond set, $\setL_S$ also d-separates some disjoint partition of measures ${\setX_i \subset \setX'}$ from ${\setX_j \subset \setX'}$, and accordingly, denote by $\graph_i$ the subgraph that contains measures $\setX_i$ and by $\graph_j$ the subgraph that contains measures $\setX_j$. Moreover, denote by $\setL_i \subset \setL_S$ the set of latent variables that d-separates $\setX_i$ from other measures.  The proof contains three steps.

In step 1, we show that for each $\setL'$'s sibling, its measured pure descendants are the same set of measured variables as that in the subgraph $\graph_i$ for some $i$. To this end, we first show that variables in each subgraph $\graph_i$ only formed clusters with variables in the same $\graph_i$. Without loss of generality, suppose there are only two subgraphs. 
Suppose for contradiction that we discovered a cluster of variables $\setV$ such that $\measuredp(\setV)$ comprises measures from $\graph_1, \graph_2$. Then $\setV$ can be separated into variable sets $\setV_1, \setV_2$ belonging to $\graph_1$, $\graph_2$ respectively. However, the minimal d-separating set for $\setV_1, \setV_2$ must not overlap, since they are in different subgraphs. This implies that either $\setV_1, \setV_2$ forms a rank deficient set by itself, which should have been discovered earlier. Hence, we reach a contradiction.

We next show that for each $\setL'$'s sibling, its measured pure descendants are not a proper subset of the measured variables as that in $\graph_i$. If it was the case, the variables in $\graph_i$ would continue to form variables with one another until $\setL_i$ is in the minimal d-separating set, where $\measuredp(\setA') = \setX_i$, hence showing the claim.

In step 2, we show that the cardinality of each sibling of $\setL'$ is equal to $|\setL_i|$ for some $i$. 
We first show that for a variable set $\mathbf{V}_i$ such that $\mathcal{M}(\mathbf{V}_i) \subseteq \mathcal{M}(\graph_i)$, it is not possible that $|\mathbf{V}_i| < |\setL_i|$. This is because if this was the case, it implies that there exists a latent set of smaller cardinality that separates the measured variables in $\graph_i$ from the rest of the graph, which contradicts the fact that $\setL_i$ is minimal. So it is always the case that $|\mathbf{V_i}| \geq |\setL_i|$.

Next, we show that as long as any such variable set $|\mathbf{V}_i| > |\setL_i|$, it will be able to form a cover with cardinality $< |\mathbf{V}_i|$. This is because if no rank deficient sets exist for cardinality $k < |\setL_i|$, $\{ \mathbf{V}_i: \text{the measured variables in } \mathbf{V}_i = \text{the measured variables in } \graph_i \}$ will form a cluster of cardinality $|\setL_i|$, since $|\mathbf{L}| > |\setL_i| < |\mathbf{V_i}|$.

Thus, by combining step 1 and step 2, we have shown that each sibling of $\setL'$ corresponding to $\setL_i$ for some $i$.

Finally, in step 3, we show that, for $\setL'$, there exist a set of siblings $\setS$, a set of children $\setC$, and a set of grandparent $\setP$, so that their union $\setS \cup \setC \cup \setP$ will not form a bond cover. Note that the measured pure descendants of siblings or grandparents are $\setX_i$, which is the reason why we need to consider the siblings and grandparents. Moreover, note that the reason $\setL'$ is formed as bond cover is that when it is formed, its slibings have not been found yet. Intuitively, now its siblings have been found, so we can find the correct clusters.

Suppose for contradiction that refining clusters will discover a new bond cover $\setL_{bond}$, and without loss of generality, suppose the minimal d-separating set involves some distinct covers $\setL_A, \setL_B$. Each of $\setL_A, \setL_B$ must respectively d-separate some partition of variables $\setV_A,\setV_B \subset \setV$ from all other variables remaining in $\setV$. We also know that in order to find rank deficiency, $||\setL_A|| + ||\setL_B|| < ||\setV_A|| + ||\setV_B||$, implying that $||\setL_A|| < ||\setV_A||$ or 
$||\setL_B|| < ||\setV_B||$. However, since $\setL_A,\setL_B$ d-separates $\setV_A, \setV_B$ from all other variables respectively, testing either $\setV_A, \setV_B$ must have been rank deficient. Since either of them are over a smaller latent cardinality ($||\setL_A||, ||\setL_B|| < ||\setL_A|| + ||\setL_B||$), one of them must have been discovered as a cluster earlier. Hence, we reach a contradiction.
\end{proof}

\subsection{Proof of Lemma 5}
\begin{customlem}{5}[Rank Invariance]
Denote by $\graphp$ the output from \textit{findCausalClusters} and by $\setL$ a latent atomic cover in $\graphp$. Then the rank constraints over $\measures$ prescribed by $\graphp$ before and after the operation $makeRoot(\setL)$ are identical. 
\end{customlem}

\begin{proof}
For an IL$^2$H graph $\graphp$ and a latent atomic cover $\setL$ in $\graphp$, after applying the makeRoot operator to $\setL$, which results in $\graph''$,  $\graphp$ and $\graph''$ are in the same Markov equivalence class. Therefore, $\graphp$ and $\graph''$ have the same rank constraints. 
\end{proof}

\subsection{Proof of Lemma 6}
\begin{customlem}{6}[Cross-Cover Test]
Given a set of variables $\mathcal{S}$, consider two latent atomic covers $\setL_A, \setL_B \in \mathcal{S}$, and a potential separating set $\setC = \{ \setL_{C_i} \} \subseteq \mathcal{S} \backslash \{\setL_A, \setL_B\}$. For each $\setL_{C_i}$, consider $\setC_i^A, \setC_i^B \subseteq PCh(\setL_{C_i})$ with $\setC_i^A, \setC_i^B \neq \emptyset$ and $\setC_i^A \cap \setC_i^B = \emptyset$, and denote the cardinality $k_i^A \coloneqq min(|\setL_{C_i}|, |\setC_i^A|)$, $k_i^B \coloneqq min(|\setL_{C_i}|, |\setC_i^B|)$, respectively. Then there is no edge between $\setL_A$ and $\setL_B$ if and only if there exists a separating set $\setC$ such that $\texttt{rank}(\Sigma_{\mathcal{A}, \mathcal{B}}) < min(|\setL_A| + \sum_i k_i^A, |\setL_B| + \sum_i k_i^B)$, where $\mathcal{A} = \{\setL_A, \setC_1^A, \setC_2^A, ...\}$ and $\mathcal{B} = \{\setL_B, \setC_1^B, \setC_2^B, ...\}$. In this case, we say that $\setC$ satisfies the cross-cover test of $\setL_A$ against $\setL_B$. 
\end{customlem}

\begin{proof}
We first show that if there is no edge between $\setL_A$ and $\setL_B$, then $\texttt{rank}(\Sigma_{\mathcal{A}, \mathcal{B}}) < min(|\setL_A| + \sum_i k_i^A, |\setL_B| + \sum_i k_i^B)$. 

Since there is no edge between $\setL_A$ and $\setL_B$, there exists a set $\setC = \{\setL_{C_i}\}$, so that given $\setC$, $\setL_A$ and $\setL_B$ are d-separated. Since $\setC_i^A$ and $\setC_i^B$ are the children of $\setL_{C_i}$, $\setC$ d-separates $\mathcal{A}$ from $\mathcal{B}$ as well. Then according to Theorem 7.1, $\texttt{rank}(\Sigma_{\mathcal{A}, \mathcal{B}}) = |\setC|$. Moreover, since $|\setC| < min(|\setL_A| + \sum_i k_i^A, |\setL_B| + \sum_i k_i^B)$, we have $\texttt{rank}(\Sigma_{\mathcal{A}, \mathcal{B}}) < min(|\setL_A| + \sum_i k_i^A, |\setL_B| + \sum_i k_i^B)$. 

Next we show that $\texttt{rank}(\Sigma_{\mathcal{A}, \mathcal{B}}) < min(|\setL_A| + \sum_i k_i^A, |\setL_B| + \sum_i k_i^B)$, then there is no edge between $\setL_A$ and $\setL_B$.

Now suppose that there are edges between $\setL_A$ and $\setL_B$. Then $\Sigma_{\mathcal{A}, \mathcal{B}}$ is not rank deficient; that is $\texttt{rank}(\Sigma_{\mathcal{A}, \mathcal{B}}) = min(|\setL_A| + \sum_i k_i^A, |\setL_B| + \sum_i k_i^B)$. Therefore, if $\texttt{rank}(\Sigma_{\mathcal{A}, \mathcal{B}}) < min(|\setL_A| + \sum_i k_i^A, |\setL_B| + \sum_i k_i^B)$, then there is no edge between $\setL_A$ and $\setL_B$.
\end{proof}

\subsection{Proof of Lemma 7}
\begin{customlem}{7}[V-Structure Test]
For any unshielded triangle $\setL_A-\setL_C-\setL_B$, let $\mathcal{A}, \mathcal{B}$ be the set of variables in Lemma \ref{Crosscover} such that $\Sigma_{\mathcal{A}, \mathcal{B}}$ was rank deficient.
Let $k = \texttt{rank}(\Sigma_{\mathcal{A}, \mathcal{B}})$, $k_1 = \texttt{rank}(\Sigma_{\mathcal{A}\cup \setL_C, \mathcal{B}})$, and $k_2 = \texttt{rank}(\Sigma_{\mathcal{A}, \mathcal{B}\cup \setL_C})$. Then, $\setL_A \rightarrow \setL_B \leftarrow \setL_C$ if and only if $k<\min(k_1, k_2)$.
\end{customlem}

\begin{proof}
 We first show that if $\setL_A \rightarrow \setL_C \leftarrow \setL_B$, then $k<\min(k_1, k_2)$.
 
 Since $\setL_A \rightarrow \setL_C \leftarrow \setL_B$, $\setL_C$ cannot be in the separation set of $\setL_A$ and $\setL_B$; that is, given $\setL_C$, $\setL_A$ and $\setL_B$ are d-connected. Hence, $k<k_1$ and $k<k_2$, and thus $k<\min(k_1, k_2)$.
 
 Next we show that if $k<\min(k_1, k_2)$, then $\setL_A \rightarrow \setL_C \leftarrow \setL_B$.
 
 Suppose $\setL_A, \setL_C, \setL_B$ do not form a v-structure; that is $\setL_A \rightarrow \setL_C \rightarrow \setL_B$ or $\setL_A \leftarrow \setL_C \leftarrow \setL_B$. Then $k=\min(k_1, k_2)$, since $\setL_C$ has been considered before in order to achieve rank deficiency of $\Sigma_{\mathcal{A}, \mathcal{B}}$. Therefore, if $k<\min(k_1, k_2)$, then $\setL_A \rightarrow \setL_C \leftarrow \setL_B$.
\end{proof}

\subsection{Proof of Lemma 8}
\begin{customlem}{8}
 Suppose $\graph$ is an IL$^2$H graph. The rank constraints are invariant with the \textit{minimal-graph operator} and the \textit{skeleton operator}; that is, $\graph$ and $\mathcal{O}_{skeleton}(\mathcal{O}_{min}(\graph))$ are rank equivalent.
\end{customlem}

\begin{proof}
We first show that the minimal-graph operator will not change rank deficiency constraints. Denote by $\graph_1$ and $\graph_2$ the graph before and after applying the minimal-graph operator, respectively. For every latent atomic cover
$\setL$ in $\graph_1$, since those three conditions hold, for any $\setC \subseteq \purechildrenp(\setL')$ and for any $\setS \subseteq \siblingsp(\setL')$ with $\setC, \setS \neq \emptyset$, $\texttt{rank}(\Sigma_{\mathcal{A},\mathcal{B}}) = |\setP|$, where $\mathcal{A} = \setC \cup \setS$ and $\mathcal{B} = \measuresp \backslash \measuredp(\mathcal{A})$. So, after merging $\setL$ to its parents $\setP$, the cardinality of the d-separation set between any two sets of variables does not change. Thus, according to Theorem 1, the rank constraints will not change after merging $\setL$ to its parents $\setP$.

Moreover, it is trivial to show that the skeleton operator will not change rank deficiency constraints, because the d-separation set between any two sets of variables will not change.  
\end{proof}

\subsection{Proof of Theorem 9}
\begin{customthm}{9}
Suppose $\graph$ is an IL$^2$H graph with measured variables $\measures$. Phases I-II in Algorithm 1 over $\measures$ can asymptotically identify the latent atomic covers of $\mathcal{O}_{min}(\graph)$, with the first two conditions in Condition 1.
\end{customthm}

\begin{proof}
Theorem 3 has shown that \textit{findCausalClusters} gives correct latent covers when there is no bond set. Furthermore, Theorem 4 shows that even in the presence of bond sets, refining the set over $Ch(\setL') \cup Sib(\setL') \cup Gp(\setL')$, where $\mathcal{M}_{\graphp}(\setL')$ forms a bond set, can correct the clusters. 
 
Phase II \textit{refineClusters} refines clusters over $Ch(\setL') \cup Sib(\setL') \cup Gp(\setL')$ in a breadth-first search from the root variable, and it ends after refining every latent cover in $\graphp$. Therefore, with this refining procedure, we will derive correct latent covers.
\end{proof}

\subsection{Proof of Theorem 10}
\begin{customthm}{10}
Suppose $\graph$ is an IL$^2$H graph with measured variables $\measures$. Algorithm \ref{overallAlg}, including Phases I-III, over $\measures$ can asymptotically identify the Markov equivalence class of $\mathcal{O}_{min}(\mathcal{O}_{s}(\graph))$.
\end{customthm}

\begin{proof}
Theorem 5 has shown that Phases I-II can find the correct clusters and latent atomic covers of $\mathcal{O}_{min}(\graph)$. Moreover, Lemma 2 and Lemma 3 have shown that by performing Cross-Cover Test and V-Structure Test, the skeleton and v structure among every triple of latent variables can be correctly identified. 

Phase III \textit{refineEdges} refines the edges over $\setL' \cup PCh(\setL') \cup PCh\big(PCh(\setL')\big)$ by performing Cross-Cover Test and V-Structure Test in a depth-first search from the root variable, and it ends until $\setL'$ does not have latent children. Therefore, with this refining procedure, we will derive correct skeletons and v structures.
\end{proof}


\subsection{More Explanations on Definition 3 (\textit{Effective Cardinality})}

The effective cardinality, defined in Definition 3, can be estimated with the following procedure.

\begin{algorithmic}
\STATE $j \gets 1$;\;
\STATE $\setC \gets \purechildren(\setL)$;\;
\WHILE {$j < |\setL|$}
    \STATE Find the largest subset of variables $\setC' \subseteq \setC$ such that $|\setC'| > |\parents(\setC')| = j$;\;
    \STATE Introduce a set of latents $\setL'$ with $|\setL'|=|\parents(\setC')|$;\;
    
    add $\setL'$ as new children of $\parents(\setC')$;\;
    \STATE $\setC \gets \setC \backslash \setC' \bigcup \setL'$;\;
    
    $j \mathrel{+}= 1$;\;
\ENDWHILE\;

\textit{\textbf{return}} $|\mathbf{C}|$
\end{algorithmic}

For example, for Figure 1(a), the effective cardinality of the pure children of $\{L_7,L_{8}\}$ is 3, because $|\{ X_4, X_5 \}| > |\{ L_7 \}|$ and we replace $\{ X_4, X_5 \}$ with a single latent variable $L'$, so the cardinality of the resulting children set is $|\{ X_6, X_7, L' \}| = 3$.

\subsection{More Explanations on Definition 4 (\textit{Latent Atomic Cover})}

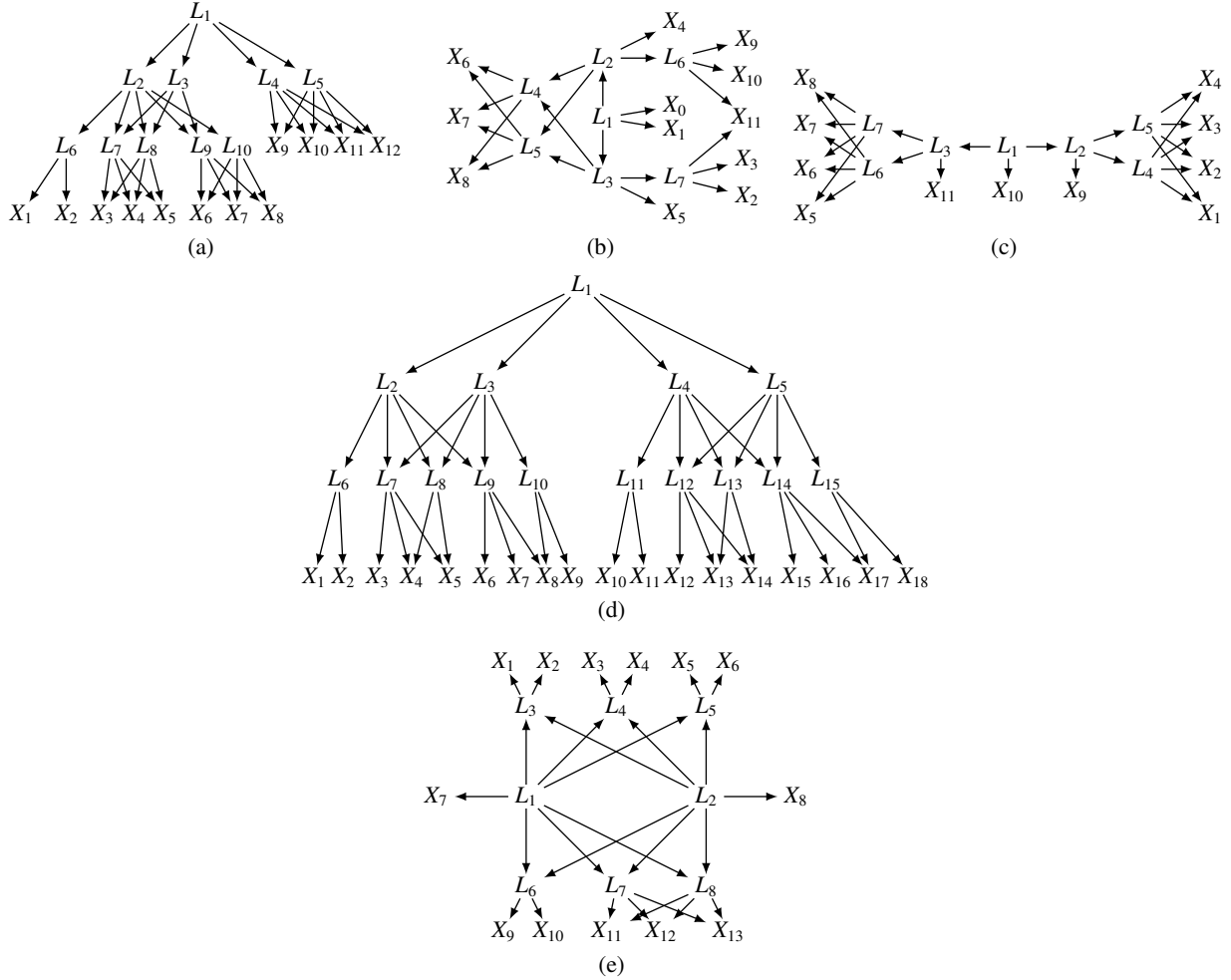
\begin{figure}[htp]
    \centering
    \subfigure[]{
	\begin{tikzpicture}[scale=.6, line width=0.5pt, inner sep=0.2mm, shorten >=.1pt, shorten <=.1pt]
		\draw (0, 2.5) node(L1)  {{\footnotesize\,$L_1$\,}};
		\draw (-1.5, 1) node(L2) {{\footnotesize\,$L_2$\,}};
		\draw (-0.5, 1) node(L3)  {{\footnotesize\,$L_3$\,}};
		\draw (1.5, 1) node(L4)  {{\footnotesize\,$L_4$\,}};
		\draw (2.5, 1) node(L5)  {{\footnotesize\,$L_5$\,}};
		
		\draw (-3, -0.5) node(L6)  {{\footnotesize\,$L_6$\,}};
		\draw (-2, -0.5) node(L7)  {{\footnotesize\,$L_7$\,}};
		\draw (-1.2, -0.5) node(L8)  {{\footnotesize\,$L_8$\,}};
		\draw (0, -0.5) node(L9)  {{\footnotesize\,$L_9$\,}};
		\draw (0.8, -0.5) node(L10)  {{\footnotesize\,$L_{10}$\,}};
		
		\draw (1.7, -0.5) node(X9)  {{\footnotesize\,$X_9$\,}};
		\draw (2.5, -0.5) node(X10)  {{\footnotesize\,$X_{10}$\,}};
		\draw (3.3, -0.5) node(X11)  {{\footnotesize\,$X_{11}$\,}};
		\draw (4.1, -0.5) node(X12)  {{\footnotesize\,$X_{12}$\,}};
		
		\draw (-4, -2) node(X1)  {{\footnotesize\,$X_1$\,}};
		\draw (-3, -2) node(X2)  {{\footnotesize\,$X_2$\,}};
		\draw (-2.2, -2) node(X3)  {{\footnotesize\,$X_3$\,}};
		\draw (-1.5, -2) node(X4)  {{\footnotesize\,$X_4$\,}};
		\draw (-0.8, -2) node(X5)  {{\footnotesize\,$X_5$\,}};
		\draw (0, -2) node(X6)  {{\footnotesize\,$X_6$\,}};
		\draw (0.8, -2) node(X7)  {{\footnotesize\,$X_7$\,}};
		\draw (1.6, -2) node(X8)  {{\footnotesize\,$X_8$\,}};
		
	   \draw[-latex] (L1) -- (L2);
	   \draw[-latex] (L1) -- (L3);
	   \draw[-latex] (L1) -- (L4);
	   \draw[-latex] (L1) -- (L5);
	   \draw[-latex] (L2) -- (L6);
	   \draw[-latex] (L2) -- (L7);
	   \draw[-latex] (L2) -- (L8);
	   \draw[-latex] (L2) -- (L9);
	   \draw[-latex] (L2) -- (L10);
	   \draw[-latex] (L3) -- (L7);
	   \draw[-latex] (L3) -- (L8);
	   \draw[-latex] (L3) -- (L9);
	   \draw[-latex] (L4) -- (X9);
	   \draw[-latex] (L4) -- (X10);
	   \draw[-latex] (L4) -- (X11);
	   \draw[-latex] (L4) -- (X12);
	   \draw[-latex] (L5) -- (X9);
	   \draw[-latex] (L5) -- (X10);
	   \draw[-latex] (L5) -- (X11);
	   \draw[-latex] (L5) -- (X12);
	   
	   \draw[-latex] (L6) -- (X1);
	   \draw[-latex] (L6) -- (X2);
	   \draw[-latex] (L7) -- (X3);
	   \draw[-latex] (L7) -- (X4);
	   \draw[-latex] (L7) -- (X5);
	   \draw[-latex] (L8) -- (X3);
	   \draw[-latex] (L8) -- (X4);
	   \draw[-latex] (L8) -- (X5);
	   \draw[-latex] (L9) -- (X6);
	   \draw[-latex] (L9) -- (X7);
	   \draw[-latex] (L9) -- (X8);
	   \draw[-latex] (L10) -- (X6);
	   \draw[-latex] (L10) -- (X7);
	   \draw[-latex] (L10) -- (X8);
	\end{tikzpicture}
    }
    \hfill
    \subfigure[]{
    \begin{tikzpicture}[scale=.8, line width=0.5pt, inner sep=0.2mm, shorten >=.1pt, shorten <=.1pt]
    \draw (0, 0) node(X6)  {{\footnotesize\,$X_6$\,}};
	\draw (0, -1) node(X7) {{\footnotesize\,$X_7$\,}};
	\draw (0, -2) node(X8)  {{\footnotesize\,$X_8$\,}};
	
	\draw (1.2, -0.5) node(L4)  {{\footnotesize\,$L_4$\,}};
	\draw (1.2, -1.5) node(L5)  {{\footnotesize\,$L_5$\,}};
	
	\draw (2.4, 0) node(L2)  {{\footnotesize\,$L_2$\,}};
	\draw (2.4, -1) node(L1) {{\footnotesize\,$L_1$\,}};
	\draw (2.4, -2) node(L3)  {{\footnotesize\,$L_3$\,}};
	\draw (3.6, 0.6) node(X4)  {{\footnotesize\,$X_4$\,}};
	\draw (3.6, -2.6) node(X5) {{\footnotesize\,$X_5$\,}};
	
	\draw (3.6, 0) node(L6)  {{\footnotesize\,$L_6$\,}};
	\draw (3.6, -0.8) node(X0) {{\footnotesize\,$X_0$\,}};
	\draw (3.6, -1.2) node(X1) {{\footnotesize\,$X_1$\,}};
	\draw (3.6, -2) node(L7)  {{\footnotesize\,$L_7$\,}};
	
	\draw (4.8, 0.3) node(X9)  {{\footnotesize\,$X_9$\,}};
	\draw (4.8, -0.3) node(X10)  {{\footnotesize\,$X_{10}$\,}};
	\draw (4.8, -1) node(X11)  {{\footnotesize\,$X_{11}$\,}};
	\draw (4.8, -1.7) node(X3)  {{\footnotesize\,$X_{3}$\,}};
	\draw (4.8, -2.3) node(X2)  {{\footnotesize\,$X_{2}$\,}};
	
	\draw[-latex] (L4) -- (X6);
	\draw[-latex] (L4) -- (X7);
	\draw[-latex] (L4) -- (X8);
	\draw[-latex] (L5) -- (X6);
	\draw[-latex] (L5) -- (X7);
	\draw[-latex] (L5) -- (X8);
	
	\draw[-latex] (L2) -- (L4);
	\draw[-latex] (L2) -- (L5);
	\draw[-latex] (L3) -- (L4);
	\draw[-latex] (L3) -- (L5);
	
	\draw[-latex] (L2) -- (X4);
	\draw[-latex] (L1) -- (L2);
	\draw[-latex] (L1) -- (L3);
	\draw[-latex] (L3) -- (X5);
	
	\draw[-latex] (L2) -- (L6);
	\draw[-latex] (L1) -- (X0);
	\draw[-latex] (L1) -- (X1);
	\draw[-latex] (L3) -- (L7);
	
	\draw[-latex] (L6) -- (X9);
	\draw[-latex] (L6) -- (X10);
	\draw[-latex] (L6) -- (X11);
	\draw[-latex] (L7) -- (X11);
	\draw[-latex] (L7) -- (X3);
	\draw[-latex] (L7) -- (X2);
	\end{tikzpicture}
    }
    \subfigure[]{
    \begin{tikzpicture}[scale=.6, line width=0.5pt, inner sep=0.2mm, shorten >=.1pt, shorten <=.1pt]
		\draw (0, 2) node(L1)  {{\footnotesize\,$L_1$\,}};
		\draw (1.5, 2) node(L2)  {{\footnotesize\,$L_2$\,}};
		\draw (-1.5, 2) node(L3) {{\footnotesize\,$L_3$\,}};
		\draw (3, 1.5) node(L4)  {{\footnotesize\,$L_4$\,}};
		\draw (3, 2.5) node(L5)  {{\footnotesize\,$L_5$\,}};
		\draw (-3, 1.5) node(L6) {{\footnotesize\,$L_6$\,}};
		\draw (-3, 2.5) node(L7) {{\footnotesize\,$L_7$\,}};
		\draw (4.5, 0.5) node(X1) {{\footnotesize\,$X_1$\,}};
		\draw (4.5, 1.5) node(X2) {{\footnotesize\,$X_2$\,}};
		\draw (4.5, 2.5) node(X3) {{\footnotesize\,$X_3$\,}};
		\draw (4.5, 3.5) node(X4) {{\footnotesize\,$X_4$\,}};
		\draw (-4.5, 0.5) node(X5) {{\footnotesize\,$X_5$\,}};
		\draw (-4.5, 1.5) node(X6) {{\footnotesize\,$X_6$\,}};
		\draw (-4.5, 2.5) node(X7) {{\footnotesize\,$X_7$\,}};
		\draw (-4.5, 3.5) node(X8) {{\footnotesize\,$X_8$\,}};
		\draw (1.5, 1) node(X9) {{\footnotesize\,$X_9$\,}};
		\draw (0, 1) node(X10) {{\footnotesize\,$X_{10}$\,}};
		\draw (-1.5, 1) node(X11) {{\footnotesize\,$X_{11}$\,}};
		
		\draw[-latex] (L1) -- (L2);
		\draw[-latex] (L1) -- (L3);
		\draw[-latex] (L2) -- (L4);
		\draw[-latex] (L2) -- (L5);
		\draw[-latex] (L3) -- (L6);
		\draw[-latex] (L3) -- (L7);
		\draw[-latex] (L2) -- (X9);
		\draw[-latex] (L1) -- (X10);
		\draw[-latex] (L3) -- (X11);
		\draw[-latex] (L4) -- (X1);
		\draw[-latex] (L4) -- (X2);
		\draw[-latex] (L4) -- (X3);
		\draw[-latex] (L4) -- (X4);
		\draw[-latex] (L5) -- (X1);
		\draw[-latex] (L5) -- (X2);
		\draw[-latex] (L5) -- (X3);
		\draw[-latex] (L5) -- (X4);
		\draw[-latex] (L6) -- (X5);
		\draw[-latex] (L6) -- (X6);
		\draw[-latex] (L6) -- (X7);
		\draw[-latex] (L6) -- (X8);
		\draw[-latex] (L7) -- (X5);
		\draw[-latex] (L7) -- (X6);
		\draw[-latex] (L7) -- (X7);
		\draw[-latex] (L7) -- (X8);
	\end{tikzpicture}
    }
    \subfigure[]{
    \begin{tikzpicture}[scale=.65, line width=0.5pt, inner sep=0.2mm, shorten >=.1pt, shorten <=.1pt]
		\draw (0, 2) node(L1)  {{\footnotesize\,$L_1$\,}};
		
		\draw (-4, 0) node(L2)  {{\footnotesize\,$L_2$\,}};
		\draw (-2, 0) node(L3) {{\footnotesize\,$L_3$\,}};
		\draw (2, 0) node(L4)  {{\footnotesize\,$L_4$\,}};
		\draw (4, 0) node(L5) {{\footnotesize\,$L_5$\,}};
		
		\draw (-5, -2) node(L6) {{\footnotesize\,$L_6$\,}};
		\draw (-4, -2) node(L7) {{\footnotesize\,$L_7$\,}};
		\draw (-3, -2) node(L8) {{\footnotesize\,$L_8$\,}};
		\draw (-2, -2) node(L9) {{\footnotesize\,$L_9$\,}};
		\draw (-1, -2) node(L10) {{\footnotesize\,$L_{10}$\,}};
		
		\draw (1, -2) node(L11) {{\footnotesize\,$L_{11}$\,}};
		\draw (2, -2) node(L12) {{\footnotesize\,$L_{12}$\,}};
		\draw (3, -2) node(L13) {{\footnotesize\,$L_{13}$\,}};
		\draw (4, -2) node(L14) {{\footnotesize\,$L_{14}$\,}};
		\draw (5, -2) node(L15) {{\footnotesize\,$L_{15}$\,}};
		
		\draw (-5.5, -4) node(X1) {{\footnotesize\,$X_1$\,}};
		\draw (-4.9, -4) node(X2) {{\footnotesize\,$X_2$\,}};
		\draw (-4.2, -4) node(X3) {{\footnotesize\,$X_3$\,}};
		\draw (-3.5, -4) node(X4) {{\footnotesize\,$X_4$\,}};
		\draw (-2.7, -4) node(X5) {{\footnotesize\,$X_5$\,}};
		\draw (-2, -4) node(X6) {{\footnotesize\,$X_6$\,}};
		\draw (-1.3, -4) node(X7) {{\footnotesize\,$X_7$\,}};
		\draw (-0.7, -4) node(X8) {{\footnotesize\,$X_8$\,}};
		\draw (-0.2, -4) node(X9) {{\footnotesize\,$X_9$\,}};
		
		\draw (0.6, -4) node(X10) {{\footnotesize\,$X_{10}$\,}};
		\draw (1.3, -4) node(X11) {{\footnotesize\,$X_{11}$\,}};
		\draw (2, -4) node(X12) {{\footnotesize\,$X_{12}$\,}};
		\draw (2.8, -4) node(X13) {{\footnotesize\,$X_{13}$\,}};
		\draw (3.6, -4) node(X14) {{\footnotesize\,$X_{14}$\,}};
		\draw (4.4, -4) node(X15) {{\footnotesize\,$X_{15}$\,}};
		\draw (5.2, -4) node(X16) {{\footnotesize\,$X_{16}$\,}};
		\draw (6.0, -4) node(X17) {{\footnotesize\,$X_{17}$\,}};
		\draw (6.8, -4) node(X18) {{\footnotesize\,$X_{18}$\,}};
		
		\draw[-latex] (L1) -- (L2);
		\draw[-latex] (L1) -- (L3);
		\draw[-latex] (L1) -- (L4);
		\draw[-latex] (L1) -- (L5);
		\draw[-latex] (L2) -- (L6);
		\draw[-latex] (L2) -- (L7);
		\draw[-latex] (L2) -- (L8);
		\draw[-latex] (L2) -- (L9);
		\draw[-latex] (L3) -- (L7);
		\draw[-latex] (L3) -- (L8);
		\draw[-latex] (L3) -- (L9);
		\draw[-latex] (L3) -- (L10);
		
		\draw[-latex] (L4) -- (L11);
		\draw[-latex] (L4) -- (L12);
		\draw[-latex] (L4) -- (L13);
		\draw[-latex] (L4) -- (L14);
		\draw[-latex] (L5) -- (L12);
		\draw[-latex] (L5) -- (L13);
		\draw[-latex] (L5) -- (L14);
		\draw[-latex] (L5) -- (L15);
		
		\draw[-latex] (L6) -- (X1);
		\draw[-latex] (L6) -- (X2);
		\draw[-latex] (L7) -- (X3);
		\draw[-latex] (L7) -- (X4);
		\draw[-latex] (L7) -- (X5);
		\draw[-latex] (L8) -- (X4);
		\draw[-latex] (L8) -- (X5);
		\draw[-latex] (L9) -- (X6);
		\draw[-latex] (L9) -- (X7);
		\draw[-latex] (L9) -- (X8);
		\draw[-latex] (L10) -- (X8);
		\draw[-latex] (L10) -- (X9);
		
		\draw[-latex] (L11) -- (X10);
		\draw[-latex] (L11) -- (X11);
		\draw[-latex] (L12) -- (X12);
		\draw[-latex] (L12) -- (X13);
		\draw[-latex] (L12) -- (X14);
		\draw[-latex] (L13) -- (X13);
		\draw[-latex] (L13) -- (X14);
		\draw[-latex] (L14) -- (X15);
		\draw[-latex] (L14) -- (X16);
		\draw[-latex] (L14) -- (X17);
		\draw[-latex] (L15) -- (X17);
		\draw[-latex] (L15) -- (X18);
	\end{tikzpicture}
    }
    
    \subfigure[]{
    \begin{tikzpicture}[scale=.6, line width=0.5pt, inner sep=0.2mm, shorten >=.1pt, shorten <=.1pt]
		\draw (0, 0) node(L1)  {{\footnotesize\,$L_1$\,}};
		\draw (4, 0) node(L2)  {{\footnotesize\,$L_2$\,}};
		
		\draw (0, 2) node(L3) {{\footnotesize\,$L_3$\,}};
		\draw (2, 2) node(L4)  {{\footnotesize\,$L_4$\,}};
		\draw (4, 2) node(L5) {{\footnotesize\,$L_5$\,}};
		
		\draw (0, -2) node(L6) {{\footnotesize\,$L_6$\,}};
		\draw (2, -2) node(L7)  {{\footnotesize\,$L_7$\,}};
		\draw (4, -2) node(L8) {{\footnotesize\,$L_8$\,}};
		
		\draw (-0.5, 3) node(X1) {{\footnotesize\,$X_1$\,}};
		\draw (0.5, 3) node(X2) {{\footnotesize\,$X_2$\,}};
		\draw (1.5, 3) node(X3) {{\footnotesize\,$X_3$\,}};
		\draw (2.5, 3) node(X4) {{\footnotesize\,$X_4$\,}};
		\draw (3.5, 3) node(X5) {{\footnotesize\,$X_5$\,}};
		\draw (4.5, 3) node(X6) {{\footnotesize\,$X_6$\,}};
		
		\draw (-2, 0) node(X7) {{\footnotesize\,$X_7$\,}};
		\draw (6, 0) node(X8) {{\footnotesize\,$X_8$\,}};
		
		\draw (-0.5, -3) node(X9) {{\footnotesize\,$X_9$\,}};
		\draw (0.5, -3) node(X10) {{\footnotesize\,$X_{10}$\,}};
		\draw (1.8, -3) node(X11) {{\footnotesize\,$X_{11}$\,}};
		\draw (3, -3) node(X12) {{\footnotesize\,$X_{12}$\,}};
		\draw (4.5, -3) node(X13) {{\footnotesize\,$X_{13}$\,}};
		
		\draw[-latex] (L3) -- (X1);
		\draw[-latex] (L3) -- (X2);
		\draw[-latex] (L4) -- (X3);
		\draw[-latex] (L4) -- (X4);
		\draw[-latex] (L5) -- (X5);
		\draw[-latex] (L5) -- (X6);
		\draw[-latex] (L1) -- (X7);
		\draw[-latex] (L2) -- (X8);
		\draw[-latex] (L6) -- (X9);
		\draw[-latex] (L6) -- (X10);
		\draw[-latex] (L7) -- (X11);
		\draw[-latex] (L7) -- (X12);
		\draw[-latex] (L7) -- (X13);
		\draw[-latex] (L8) -- (X11);
		\draw[-latex] (L8) -- (X12);
		\draw[-latex] (L8) -- (X13);
		\draw[-latex] (L1) -- (L3);
		\draw[-latex] (L1) -- (L4);
		\draw[-latex] (L1) -- (L5);
		\draw[-latex] (L2) -- (L3);
		\draw[-latex] (L2) -- (L4);
		\draw[-latex] (L2) -- (L5);
		\draw[-latex] (L1) -- (L6);
		\draw[-latex] (L1) -- (L7);
		\draw[-latex] (L1) -- (L8);
		\draw[-latex] (L2) -- (L6);
		\draw[-latex] (L2) -- (L7);
		\draw[-latex] (L2) -- (L8);
	\end{tikzpicture}
    }
	\caption{Latent hierarchical graphs where some latent atomic covers only have $k$ neighbors, except for the $k+1$ pure children, but the graph structure is still identifiable. Note that in many cases, $k$-neighbor is enough; the condition ``$k+1$ neighbors" is sufficient, but not necessary.}
	\label{fig:not IL2H}
\end{figure}

The first two conditions in Definition 4 ensure that there are enough variables in the current active variable set to find the rank deficiency, so that we can determine the latent atomic cover with size $k$. However, note that they may not be the necessary conditions. For example, for the graphs in Figure \ref{fig:not IL2H}, some of the latent atomic covers only have $k$ neighbors (except for the $k+1$ pure children), but they are still identifiable.

The first half of the third condition, ``there does not exist a partition of $\setL = \setL_1 \cup \setL_2$, so that both $\setL_1,\setL_2$ satisfy conditions 1 and 2", ensure that the latent atomic cover $\setL$ is atomic.

The second half of the third condition, ``there does not exist a partition of $\setL = \setL_1 \cup \setL_2$, so that $\{PCh_{\graph}(\setL_1) \cup PCh_{\graph}(\setL_2)\} \backslash \setL = PCh_{\graph}(\setL)$", is needed in the overlapping case. Consider the following graph: $L_1 \rightarrow \{X_1,X_2,X_3\}, L_2 \rightarrow \{X_3,X_4,X_5\}$. Here, $\setL_1 = \{L_1\}$, $\setL_2 = \{L_2\}$, and $\setL = \{L_1,L_2\}$ are latent atomic covers; however, $\{PCh_{\graph}(\setL_1) = \{X_1,X_2\}$, $\{PCh_{\graph}(\setL_2) = \{X_4,X_5\}$, and $\{PCh_{\graph}(\setL) = \{X_1,X_2,X_3,X_4,X_5\}$, so $\{PCh_{\graph}(\setL_1) \cup PCh_{\graph}(\setL_2)\} \backslash \setL \neq PCh_{\graph}(\setL)$. Therefore, although both $\setL_1$ and $\setL_2$ satisfy the first two conditions, $\setL$ satisfies Definition 4.

\subsection{More Explanations on Condition 1 (\textit{IL$^2$H graph})}
The first two conditions in Condition 1 guarantee to identify the latent atomic covers, while the last one is used to identify the edges among latent atomic covers when performing Cross-Cover Test and V-Structure Test. 

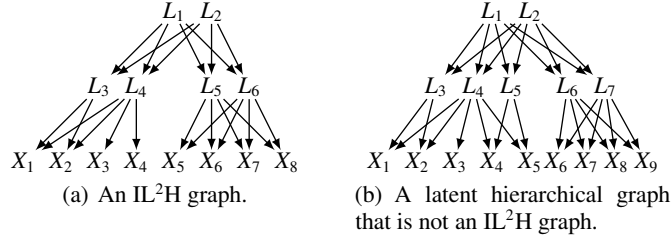
\begin{figure}[htp!]
    \centering
    \subfigure[An IL$^2$H graph.]{
    \begin{tikzpicture}[scale=.5, line width=0.5pt, inner sep=0.2mm, shorten >=.1pt, shorten <=.1pt]
		\draw (2, 4) node(L1)  {{\footnotesize\,$L_1$\,}};
		\draw (3, 4) node(L2)  {{\footnotesize\,$L_2$\,}};
		\draw (0, 2) node(L3) {{\footnotesize\,$L_3$\,}};
		\draw (1, 2) node(L4) {{\footnotesize\,$L_4$\,}};
		\draw (3, 2) node(L5) {{\footnotesize\,$L_5$\,}};
		\draw (4, 2) node(L6) {{\footnotesize\,$L_6$\,}};
		\draw (-2, 0) node(X1)  {{\footnotesize\,$X_1$\,}};
		\draw (-1, 0) node(X2)  {{\footnotesize\,$X_2$\,}};
		\draw (0, 0) node(X3) {{\footnotesize\,$X_3$\,}};
		\draw (1, 0) node(X4) {{\footnotesize\,$X_4$\,}};
		\draw (2, 0) node(X5)  {{\footnotesize\,$X_5$\,}};
		\draw (3, 0) node(X6)  {{\footnotesize\,$X_6$\,}};
		\draw (4, 0) node(X7) {{\footnotesize\,$X_7$\,}};
		\draw (5, 0) node(X8) {{\footnotesize\,$X_8$\,}};
		
		\draw[-latex] (L1) -- (L3); 
		\draw[-latex] (L1) -- (L4);
		\draw[-latex] (L1) -- (L5); 
		\draw[-latex] (L1) -- (L6);
		\draw[-latex] (L2) -- (L3); 
		\draw[-latex] (L2) -- (L4);
		\draw[-latex] (L2) -- (L5); 
		\draw[-latex] (L2) -- (L6);
		\draw[-latex] (L3) -- (X1);
		\draw[-latex] (L3) -- (X2);
		\draw[-latex] (L4) -- (X1);
		\draw[-latex] (L4) -- (X2);
		\draw[-latex] (L4) -- (X3);
		\draw[-latex] (L4) -- (X4);
		\draw[-latex] (L5) -- (X5);
		\draw[-latex] (L5) -- (X6);
		\draw[-latex] (L5) -- (X7);
		\draw[-latex] (L5) -- (X8);
		\draw[-latex] (L6) -- (X5);
		\draw[-latex] (L6) -- (X6);
		\draw[-latex] (L6) -- (X7);
		\draw[-latex] (L6) -- (X8);
	\end{tikzpicture}
	\label{example-2a}
    }
    \hspace{0.5cm}%
    \subfigure[A latent hierarchical graph that is not an IL$^2$H graph.]{
    \begin{tikzpicture}[scale=.5, line width=0.5pt, inner sep=0.2mm, shorten >=.1pt, shorten <=.1pt]
		\draw (1.5, 4) node(L1)  {{\footnotesize\,$L_1$\,}};
		\draw (2.5, 4) node(L2)  {{\footnotesize\,$L_2$\,}};
		\draw (0, 2) node(L3) {{\footnotesize\,$L_3$\,}};
		\draw (1, 2) node(L4)  {{\footnotesize\,$L_4$\,}};
		\draw (2, 2) node(L5)  {{\footnotesize\,$L_5$\,}};
		\draw (3.5, 2) node(L6) {{\footnotesize\,$L_6$\,}};
		\draw (4.5, 2) node(L7) {{\footnotesize\,$L_7$\,}};
		\draw (-1.5, 0) node(X1) {{\footnotesize\,$X_1$\,}};
		\draw (-0.5, 0) node(X2) {{\footnotesize\,$X_2$\,}};
		\draw (0.5, 0) node(X3) {{\footnotesize\,$X_3$\,}};
		\draw (1.5, 0) node(X4) {{\footnotesize\,$X_4$\,}};
		\draw (2.5, 0) node(X5) {{\footnotesize\,$X_5$\,}};
		\draw (3.2, 0) node(X6) {{\footnotesize\,$X_6$\,}};
		\draw (4, 0) node(X7) {{\footnotesize\,$X_7$\,}};
		\draw (4.8, 0) node(X8) {{\footnotesize\,$X_8$\,}};
		\draw (5.6, 0) node(X9) {{\footnotesize\,$X_9$\,}};
		\draw[-latex] (L1) -- (L3); 
		\draw[-latex] (L1) -- (L4); 
		\draw[-latex] (L1) -- (L5); 
		\draw[-latex] (L1) -- (L6); 
		\draw[-latex] (L1) -- (L7); 
		\draw[-latex] (L2) -- (L3); 
		\draw[-latex] (L2) -- (L4); 
		\draw[-latex] (L2) -- (L5); 
		\draw[-latex] (L2) -- (L6); 
		\draw[-latex] (L2) -- (L7); 
		\draw[-latex] (L3) -- (X1); 
		\draw[-latex] (L3) -- (X2);
		\draw[-latex] (L4) -- (X1);
		\draw[-latex] (L4) -- (X2);
		\draw[-latex] (L4) -- (X3);
		\draw[-latex] (L4) -- (X4);
		\draw[-latex] (L4) -- (X5);
		\draw[-latex] (L5) -- (X4);
		\draw[-latex] (L5) -- (X5);
		\draw[-latex] (L6) -- (X6);
		\draw[-latex] (L6) -- (X7);
		\draw[-latex] (L6) -- (X8);
		\draw[-latex] (L6) -- (X9);
		\draw[-latex] (L7) -- (X6);
		\draw[-latex] (L7) -- (X7);
		\draw[-latex] (L7) -- (X8);
		\draw[-latex] (L7) -- (X9);
	\end{tikzpicture}
	\label{example-2c}
    }
    \caption{Examples and counter-examples of IL$^2$H graphs.}
    \label{example-2}
\end{figure}

Specifically, the second condition says that two latent atomic covers that are partly overlapped is not allowed, except for the case that one latent atomic cover is contained in another one; otherwise, the cover creation rule can be non-trivial.  
For example, for the graph in Figure \ref{example-2a}, although $ \puredescendant(\{ L_3, L_4 \}) \bigcap \puredescendant(\{L_4\}) = \{X_3, X_4\}$, it satisfies the second condition in Condition 1 because $\{L_3\} \subset \{ L_2, L_3 \}$. In contrast, for the graph in Figure \ref{example-2c}, $\puredescendant(\{L_3, L_4\}) \bigcap \puredescendant(\{L_4, L_5\}) = \{X_3\}$, but $\{L_3, L_4\}, \{L_4, L_5\}$ are not subsets or descendants of one another. Hence, it does not satisfy the second condition of an IL$^2$H graph.


\subsection{More Explanations on Algorithm 2 (\textit{findCausalClusters})}

The search procedure in Algorithm 2 (\textit{findCausalClusters}) contains the following two key updates.
\begin{itemize}[leftmargin=10pt,itemsep=0pt,topsep=-1pt]
    \item \textit{Latent-atomic-cover size update.} We start to identify latent atomic covers with size $k=1$. If a rank-deficiency set is not found, then increment $k=k+1$; otherwise, reset $k=1$.
    \item \textit{Active variable set update.} The set of active variables $\mathcal{S}$ is set to $\measures$ initially. We consider any subset of the latent atomic covers in $\mathcal{S}$ and replace them with their pure children, resulting in $\tilde{\mathcal{S}}$. 
    At the same time, we search for the rank-deficiency set $\setA$ with rank $k$ in $\tilde{\mathcal{S}}$; if it is found, then assign a latent atomic cover $\setL$ of size $k$ as the parent $\setA$, and accordingly, the active variable set is updated as $(\mathcal{S} \backslash \setA) \cup \setL$. 
\end{itemize}

\begin{figure}[hpt!]
    \centering
    \subfigure[]{
	\begin{tikzpicture}[scale=.6, line width=0.5pt, inner sep=0.1mm, shorten >=.1pt, shorten <=.1pt]
		\draw (-0.2, 0) node(X6)  {{\footnotesize\,$X_6$\,}};
		\draw (0.6, 0) node(X7) {{\footnotesize\,$X_7$\,}};
		\draw (1.4, 0) node(X8) {{\footnotesize\,$X_8$\,}};
		\draw (2.2, 0) node(X9) {{\footnotesize\,$X_9$\,}};
		\draw (3, 0) node(X2) {{\footnotesize\,$X_2$\,}};
		\draw (4, 0) node(X3) {{\footnotesize\,$X_3$\,}};
		\draw (5, 0) node(X12) {{\footnotesize\,$X_{12}$\,}};
		\draw (6, 0) node(X10) {{\footnotesize\,$X_{10}$\,}};
		\draw (7, 0) node(X11) {{\footnotesize\,$X_{11}$\,}};
		\draw (0.5, 1.5) node(L4) {{\footnotesize\,$L_4'$\,}};
		\draw (1.5, 1.5) node(L5) {{\footnotesize\,$L_5'$\,}};
		\draw (3.5, 1.5) node(L7) {{\footnotesize\,$L_7'$\,}};
		\draw (6.5, 1.5) node(L6) {{\footnotesize\,$L_6'$\,}};
		\draw (2, 3) node(L3) {{\footnotesize\,$L_3'$\,}};
		\draw (4, 3) node(L2) {{\footnotesize\,$L_2'$\,}};
		\draw (0, 2.5) node(X5) {{\footnotesize\,$X_5$\,}};
		\draw (6, 2.5) node(X4) {{\footnotesize\,$X_4$\,}};
		\draw (0.5, 5) node(L1) {{\footnotesize\,$L_1'$\,}};
		\draw (0, 3.5) node(X0) {{\footnotesize\,$X_0$\,}};
		\draw (1, 3.5) node(X1) {{\footnotesize\,$X_1$\,}};

	   \draw[-latex] (L4) -- (X6);
	   \draw[-latex] (L4) -- (X7);
	   \draw[-latex] (L4) -- (X8);
	   \draw[-latex] (L4) -- (X9);
	   \draw[-latex] (L5) -- (X6);
	   \draw[-latex] (L5) -- (X7);
	   \draw[-latex] (L5) -- (X8);
	   \draw[-latex] (L5) -- (X9);
	   \draw[-latex] (L7) -- (X2);
	   \draw[-latex] (L7) -- (X3);
	   \draw[-latex] (L7) -- (X12);
	   \draw[-latex] (L6) -- (X10);
	   \draw[-latex] (L6) -- (X11);
	   \draw[-latex] (L6) -- (X12);
	   
	    \draw[-latex] (L1) -- (L2);
	   \draw[-latex] (L3) -- (L4);
	   \draw[-latex] (L3) -- (L5);
	   \draw[-latex] (L3) -- (L7);
	   \draw[-latex] (L3) -- (X5);
	   \draw[-latex] (L2) -- (L4);
	   \draw[-latex] (L2) -- (L5);
	   \draw[-latex] (L2) -- (L6);
	   \draw[-latex] (L2) -- (X4);
	   \draw[-latex] (L1) -- (L3);
	   \draw[-latex] (L1) -- (X0);
	   \draw[-latex] (L1) -- (X1);
	   	\draw[blue,dashed] (5,0.5) ellipse (68pt and 46pt);
	\end{tikzpicture}
    }
    \hfill
    ~~~
    \subfigure[]{
	\begin{tikzpicture}[scale=.6, line width=0.5pt, inner sep=0.1mm, shorten >=.1pt, shorten <=.1pt]
		\draw (-0.2, 0) node(X6)  {{\footnotesize\,$X_6$\,}};
		\draw (0.6, 0) node(X7) {{\footnotesize\,$X_7$\,}};
		\draw (1.4, 0) node(X8) {{\footnotesize\,$X_8$\,}};
		\draw (2.2, 0) node(X9) {{\footnotesize\,$X_9$\,}};
		\draw (3, 0) node(X2) {{\footnotesize\,$X_2$\,}};
		\draw (4, 0) node(X3) {{\footnotesize\,$X_3$\,}};
		\draw (5, 0) node(X10) {{\footnotesize\,$X_{10}$\,}};
		\draw (6, 0) node(X11) {{\footnotesize\,$X_{11}$\,}};
		\draw (7, 0) node(X0) {{\footnotesize\,$X_0$\,}};
		\draw (8, 0) node(X1) {{\footnotesize\,$X_1$\,}};
		
		\draw (0.5, 1.5) node(L4) {{\footnotesize\,$L_4'$\,}};
		\draw (1.5, 1.5) node(L5) {{\footnotesize\,$L_5'$\,}};
		\draw (3.5, 1.5) node(L7) {{\footnotesize\,$L_7'$\,}};
		\draw (5.5, 1.5) node(L6) {{\footnotesize\,$L_6'$\,}};
		\draw (7.5, 1.5) node(L1) {{\footnotesize\,$L_1'$\,}};
		\draw (2.5, 1.5) node(X4) {{\footnotesize\,$X_4$\,}};
		\draw (4.5, 1.5) node(X5) {{\footnotesize\,$X_5$\,}};
		\draw (6.5, 1.5) node(X12) {{\footnotesize\,$X_{12}$\,}};
		
		\draw (3, 3) node(L8) {{\footnotesize\,$L_8'$\,}};
		\draw (5, 3) node(L9) {{\footnotesize\,$L_9'$\,}};
		
		\draw[-latex] (L4) -- (X6);
		\draw[-latex] (L4) -- (X7);
		\draw[-latex] (L4) -- (X8);
		\draw[-latex] (L4) -- (X9);
		\draw[-latex] (L5) -- (X6);
		\draw[-latex] (L5) -- (X7);
		\draw[-latex] (L5) -- (X8);
		\draw[-latex] (L5) -- (X9);
		\draw[-latex] (L7) -- (X2);
		\draw[-latex] (L7) -- (X3);
		\draw[-latex] (L6) -- (X10);
		\draw[-latex] (L6) -- (X11);
		\draw[-latex] (L1) -- (X0);
		\draw[-latex] (L1) -- (X1);
		\draw[-latex] (L8) -- (L4);
		\draw[-latex] (L8) -- (L5);
		\draw[-latex] (L8) -- (L7);
		\draw[-latex] (L8) -- (L6);
		\draw[-latex] (L8) -- (L1);
		\draw[-latex] (L8) -- (X4);
		\draw[-latex] (L8) -- (X5);
		\draw[-latex] (L8) -- (X12);
		\draw[-latex] (L9) -- (L4);
		\draw[-latex] (L9) -- (L5);
		\draw[-latex] (L9) -- (L7);
		\draw[-latex] (L9) -- (L6);
		\draw[-latex] (L9) -- (L1);
		\draw[-latex] (L9) -- (X4);
		\draw[-latex] (L9) -- (X5);
		\draw[-latex] (L9) -- (X12);
		\draw[blue,dashed] (4.8,0.8) ellipse (58pt and 38pt);
	\end{tikzpicture}
    }
\caption{(a) Output from Phase I \textit{findCausalClusters} with ground truth graph in Figure \ref{fig:IL2H}(b), where we consider the children of the active variable set $\mathcal{S}$ (i.e., $\tilde{\mathcal{S}}$) to search for the rank deficiency set. (b) The output graph if directly searching over $\mathcal{S}$ in Phase I; that is, without considering line 4 in Algorithm 2. The output graph without considering $\tilde{\mathcal{S}}$ is not correct; for instance, $X_{12}$ will not be considered as the children of $\{L_6,L_7\}$. }
\label{Counterexp: phase I}
\end{figure}
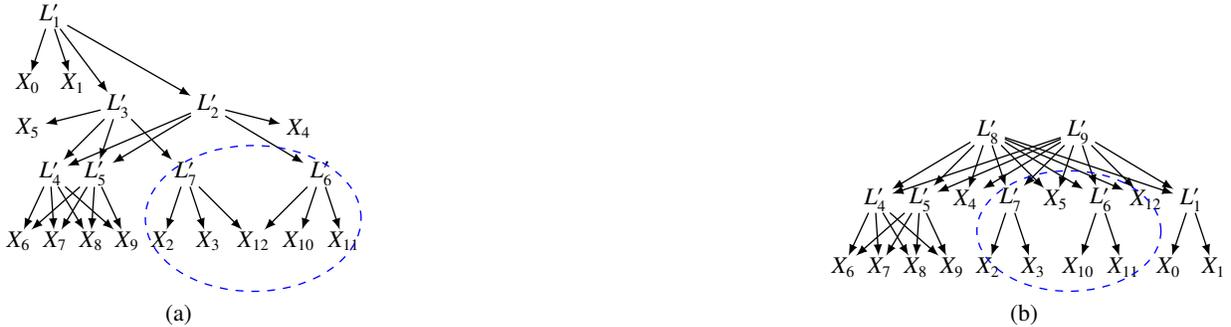

Note that in Algorithm 2 \textit{findCausalClusters}, we search over $\tilde{\mathcal{S}}$, instead of $\mathcal{S}$, to avoid adding a certain type of redundant latent variables when some latent atomic covers have overlapping variables, including the v structure; see Figure \ref{Counterexp: phase I} for an illustration.

Moreover, in Algorithm 2 \textit{findCausalClusters}, if there are conflicts when the search goes on, then we just ignore it, and such conflicts will be handled in Algorithm 3 \textit{refineClusters}. Also, note that except for the v structure where the measured variable is a collider, other ``v structures" in the intermediate output in Algorithms 2 and 3 are not true v structures.

For the illustration of Algorithm 2 given in Figure 2, here we give more detailed explanations. Specifically, we first set $k=1$ and the active set is $\mathcal{S} = \big\{ X_1,\cdots,X_{16}\big\}$ and $\tilde{\mathcal{S}} = \mathcal{S}$, and we can find the clusters in (a), and no further cluster can be found with $k=1$. Then we increase $k$ to $2$ with the active set $\mathcal{S} = \{\big\{L_6\},\{L_7\},X_6,\cdots,X_{16}\big\}$ and $\tilde{\mathcal{S}} = \mathcal{S}$, and then we can find the clusters in (b).  
Then, the active set is $\mathcal{S} = \big\{\{L_4, L_5\},\{L_6\},\{L_7,L_8\},\{L_9,L_{10}\}\big\}$ and we set back $k=1$, and when $\tilde{\mathcal{S}}=\{\{L_4,L_5\}, X_1,\cdots,X_{11}\}$ we find the cluster in (c).  Note that when testing the rank over $\{L_4,L_5\}$ against other variables, we use their measured pure descendants in the currently estimated graph instead. The above procedure is repeated to further find the cluster in (d). Finally, when there are no enough variables for testing, we connect the elements in the active variable set: connecting $\{L_2,L_3\}$ to $\{L_7,L_8\}$ in (e).




\subsection{Complexity of Algorithm 1}
The time complexity of the algorithm is upper bounded by $O(r \sum_{k=2}^{l+1}\binom{m}{k})$, and this bound is further upper bounded by $O(r(1+m)^{l+1})$, where $m$ is the number of measured variables, $l$ is the cardinality of the largest latent cover of the estimated graph, with $l \ll m$, and $r$ is the number of levels of the estimated hierarchical graph, with $r \ll m$.

\subsection{Illustrative Examples of the Entire Algorithm}

\begin{figure}[hpt!]
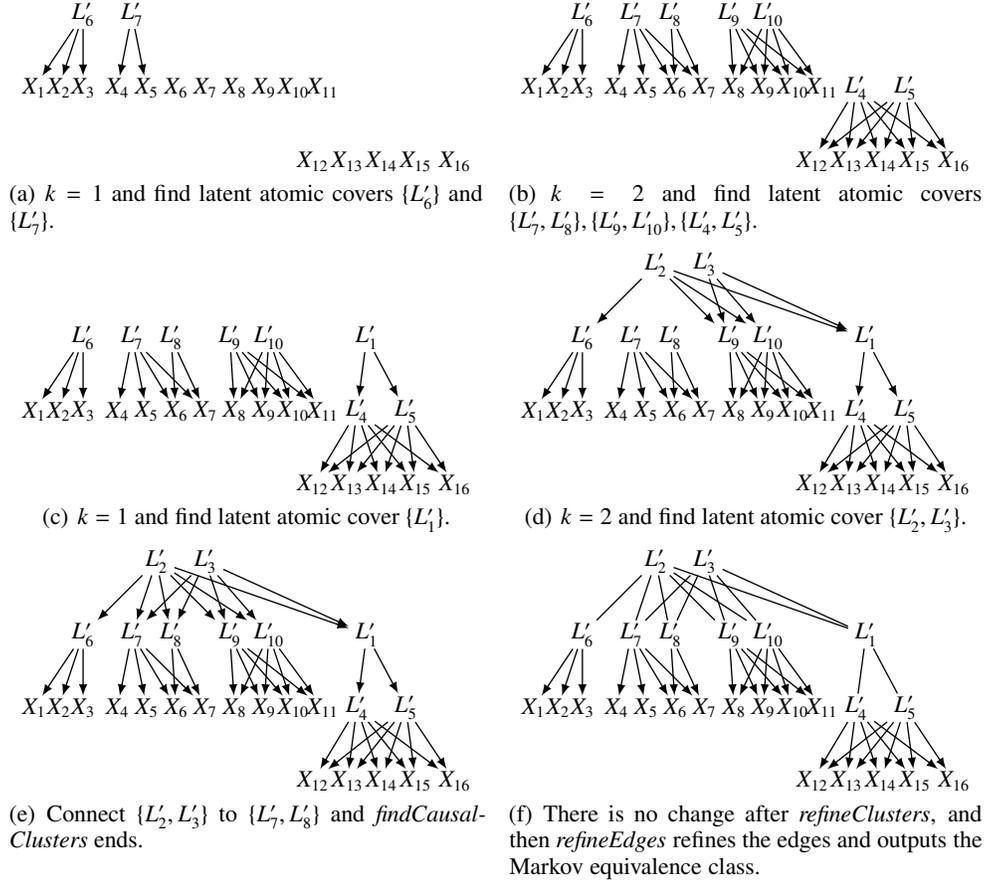

    \centering
    \subfigure[$k=1$ and find latent atomic covers $\{L_6'\}$ and $\{L_7'\}$.]{

    }
    ~~
        \subfigure[$k=2$ and find latent atomic covers $\{L_7',L_8'\}, \{L_9',L_{10}'\}, \{L_4',L_5'\}$. ]{
	%
    }
    ~~
        \subfigure[$k=1$ and find latent atomic cover $\{L_1'\}$.]{
	%
    }
    ~~
        \subfigure[$k=2$ and find latent atomic cover $\{L_2',L_3'\}$.]{
	%
    }
   ~~
     \subfigure[There is no change after \textit{refineClusters}, and then \textit{refineEdges} refines the edges and outputs the Markov equivalence class.]{
	%
    }
\caption{An illustrative example by applying Algorithm 1 to the measured variables in Figure \ref{fig:IL2H}(a).}
\label{fig:illustration 1}
\end{figure}

Figure \ref{fig:illustration 1} and Figure \ref{fig:illustration 2} give two illustrative examples of the entire algorithm, showing how each step proceeds.

\begin{figure}[htp]
    \centering
    \subfigure[$k=2$ and find latent atomic covers $\{L_6',L_7'\}$, $\{L_4',L_5'\}$, and $\{L_8',L_9'\}$.]{
    %
    }
    ~~
    \subfigure[$k=1$ and find a latent atomic cover $\{L_2'\}$.]{
    %
    }
    ~~
    \subfigure[$k=1$ and find a latent atomic cover $\{L_3'\}$.]{
    %
    }
    ~~
    \subfigure[Refine $\{L_2'\}$ by first removing $\{L_2'\}$ and its parents $\{L_8',L_9'\}$ and performing \textit{findCausalClusters} over $\{L_3',L_4',L_5',X_9,X_{10},X_{11}\}$, and then we can find a latent cover $\{L_{10}'\}$.]{
    %
    }
    ~~
    \subfigure[Next perform \textit{findCausalClusters} over $\{L_{10}',L_4',L_5',X_9,X_{10}\}$ we can find a latent cover $\{L_{11}'\}$.]{
    %
    }
    ~~
    \subfigure[Next perform \textit{findCausalClusters} over $\{L_{11}',L_4',L_5',X_9\}$ we can find a latent cover $L_{12}'$.]{
    %
    }
     ~~
    \subfigure[Refine $\{L_3'\}$ by first removing $\{L_3'\}$ and its parents $\{L_{10}'\}$ and performing \textit{findCausalClusters} over $\{L_6', L_7', L_{11}', X_{11}\}$, and then we can find a latent cover $\{L_{13}'\}$.]{
    %
    }
    ~~
    \subfigure[Perform \textit{refineEdges} to refine the edges and output the Markov equivalence class.]{
    %
    }

\setlength{\abovecaptionskip}{-4pt}
	\caption{An illustrative example by applying Algorithm 1 to the measured variables in Figure \ref{fig:not IL2H}(c).}
	\label{fig:illustration 2}
\end{figure}

\bibliography{references}
\end{document}